\pdfoutput=1
\documentclass[twoside]{article}

\usepackage[preprint]{aistats2027} %

\usepackage{amsthm, amssymb, dsfont, amsmath, bm}
\usepackage{xcolor}
\usepackage{fontawesome5}
\usepackage{graphicx}
\usepackage[english]{babel}
\usepackage{xfrac}
\usepackage{lmodern}
\usepackage{algorithm}
\usepackage{algpseudocode}
\usepackage{booktabs}
\usepackage{multirow}
\usepackage{xspace}
\usepackage{tikz}
\usepackage{dirtytalk}
\usepackage[hidelinks]{hyperref} %
\usepackage{cleveref} %

\definecolor{lvlB}{HTML}{EE7733}   %
\definecolor{lvlC}{HTML}{009988}   %
\newcommand{\Lmm}[1]{\textcolor{lvlB}{#1}}
\newcommand{\Ldink}[1]{\textcolor{lvlC}{#1}}

\theoremstyle{plain}

\theoremstyle{definition}

\usepackage[round]{natbib}

\newcommand{\x}{\mathbf{x}}
\newcommand{\z}{\mathbf{z}}
\newcommand{\uu}{\mathbf{u}}       %
\newcommand{\vv}{\mathbf{v}}       %
\newcommand{\w}{\mathbf{w}}
\newcommand{\y}{\mathbf{y}}
\newcommand{\g}{\mathbf{g}}

\newcommand{\one}{\mathbf{1}}
\newcommand{\zero}{\mathbf{0}}

\newcommand{\vphi}{\bm{\phi}}

\newcommand{\A}{\mathbf{A}}

\newcommand{\PP}{\mathbf{P}}       %
\newcommand{\Z}{\mathbf{Z}}
\newcommand{\I}{\mathbf{I}}
\newcommand{\Sig}{\bm{\Sigma}}
\newcommand{\Del}{\bm{\Delta}}
\newcommand{\Ktil}{\widetilde{\mathbf{K}}}
\newcommand{\R}{\mathbb{R}}
\newcommand{\Sphere}{\mathbb{S}}
\newcommand{\Xcal}{\mathcal{X}}
\DeclareMathOperator*{\argmax}{arg\,max}
\DeclareMathOperator*{\argmin}{arg\,min}
\definecolor{timegray}{gray}{0.5}
\newcommand{\tcell}[1]{\textcolor{timegray}{(#1)}}

\newcommand{\methodname}{\mbox{Fast BO}\xspace}

\begin{document}

\runningtitle{Fast Bayesian Optimization Through Linear Models}
\runningauthor{Colin Doumont, Henry Moss, Philipp Hennig}

\twocolumn[
  \aistatstitle{Fast Bayesian Optimization Through Linear Models}
  \aistatsauthor{Colin Doumont \And Henry Moss \And Philipp Hennig}
  \aistatsaddress{University of T\"ubingen \\ T\"ubingen AI Center \And Lancaster University \And University of T\"ubingen \\ T\"ubingen AI Center}
  \begin{center}
    \vspace{-0.22in} \normalsize \href{https://github.com/colmont/fast-bo}{\faGithub\ \textcolor{blue}{\texttt{https://github.com/colmont/fast-bo}} } \vspace{14.44pt} %
  \end{center}
]

\begin{abstract}
  Bayesian optimization (BO) is typically used only for problems where the underlying function is deemed \say{expensive} (e.g.\ in time or money), with the speed of the BO algorithm being of little importance. In the real world, however, problems span a continuous range of values, going from cheap to expensive. In this paper, we present BO algorithms through the lens of a speed-performance Pareto frontier: we compare BO against faster algorithms like \texttt{CMA-ES}, and highlight how certain regions of the Pareto frontier are currently empty. As a solution, we propose a new set of fast BO methods that exploit the linear structure in existing BO algorithms, for both continuous and combinatorial inputs (from ${\sim}10$ to $6000{+}$ dimensions). Our methods achieve performance similar to state-of-the-art BO, while being more than $10^2 \times$ to $10^3 \times$ faster, opening the field of BO to an entirely new class of problems.
\end{abstract}

\section{Introduction}

Black-box optimization, i.e.\ optimization of a function that is only observable through evaluations (e.g.\ a simulator or physical experiment), has been studied since \citet{boxs1951exp}. Over the past 75 years, different techniques and approaches have been proposed, ranging from direct search \citep{nelder1965simplex} and local surrogates \citep{powell2009bobyqa} to evolutionary algorithms \citep{holland1975adaptation, hansen2001completely}. Among these, Bayesian optimization \citep{movckus1975bayesian,jones1998efficient} became known for its sample efficiency, and is now often the default black-box optimizer for certain applications \citep[e.g.\ hyperparameter optimization in machine learning;][]{snoek2012practical}. Nevertheless, until recently, BO was still mostly constrained to low-dimensional continuous spaces (i.e.\ below 20 dimensions), with a fairly limited number of observations \citep{frazier2018bayesian}.

\begin{figure}[t]
  \centering
  \setlength{\abovecaptionskip}{3pt}
  \includegraphics[scale=0.806]{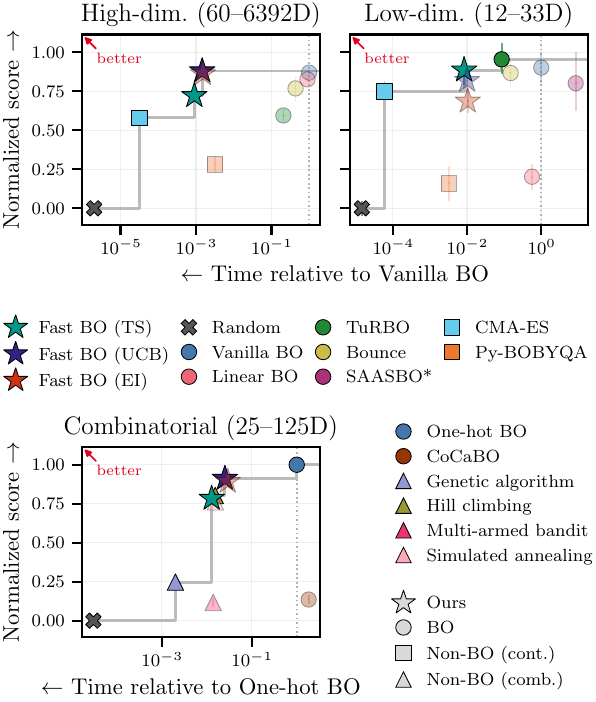}
  \caption{Summary plot of the speed-performance Pareto frontiers in \Cref{fig:pareto-cont-high,fig:pareto-cont-low,fig:pareto-comb}. Speed is measured relative to state-of-the-art BO (averaged using a geometric mean). For performance, we rescale the final best values of each benchmark such that random search scores $0$ and the best method scores $1$ (and then average these scores over all benchmarks). In all three settings, \textbf{our fast BO methods cover new areas of the Pareto frontier}, matching many well-known BO methods at a fraction of the cost. Moreover, \texttt{CMA-ES} performs surprisingly well --- matching \texttt{TuRBO} on high-dimensional problems --- highlighting the relevance of fast black-box optimizers.}
  \label{fig:pareto-aggregate}
\end{figure}

In the last decade, a growing body of literature has focused on extending the method to a larger set of problems --- such as combinatorial spaces \citep{doumont2025omnipresent} or higher dimensions \citep{hvarfner2024vanilla} --- enabling BO to spread to new application areas \citep{moss2025,fan2026search}. However, one major limiting factor is still largely unaddressed: BO methods are typically much slower than other black-box optimizers (see \Cref{fig:pareto-aggregate}). For example, current state-of-the-art BO methods can take multiple minutes to choose the next acquisition. If evaluating the black-box function is cheaper than this overhead, then the BO algorithm itself becomes the main cost \citep{shahriari2015taking}, leading practitioners to instead turn to faster but less sample-efficient black-box optimizers, such as \texttt{CMA-ES} \citep{hansen2001completely,raponi2023optimizing}.

Recently, \citet{doumont2026linear} replaced the Gaussian-process surrogate by a (spherically projected) Bayesian linear model, enabling exact BO to scale to over $20{,}000$ observations, while achieving state-of-the-art performance (on high-dimensional continuous problems). However, as demonstrated in \Cref{fig:pareto-aggregate}, their method is still as slow as standard BO methods, primarily due to gradient-based techniques used for (i) fitting hyperparameters, and (ii) optimizing acquisition functions. Instead, in this paper, we show how the structure of this spherical-linear model can be exploited for both tasks, leading to speedups of several orders of magnitude with little to no loss in performance.

Specifically, for hyperparameter optimization, we replace the per-dimension lengthscales with a single shared lengthscale. This allows us to fit the resulting model efficiently through a combination of (batched) grid search and direct search \citep{nelder1965simplex}, where we reuse a single eigendecomposition across marginal-likelihood evaluations. For acquisition optimization, we reformulate Thompson sampling \citep{thompson1933likelihood} over the input box as a quadratic fractional program, which can then be solved using Dinkelbach's algorithm \citep{dinkelbach1967nonlinear} with exact $O(D)$ subproblems. This algorithm guarantees superlinear convergence to a \emph{global optimum}. For UCB \citep{srinivas2010gaussian} and EI \citep{movckus1975bayesian}, we derive linear minorizers optimized by the same exact solver, yielding minorize--maximize algorithms with guarantees of monotonic improvement and \emph{stationarity}. Interestingly, the algorithms for UCB and EI differ only in their exploration weight, which lets us view EI as a version of UCB with self-tuning exploration.

Lastly, through a standardized framework, we show how these gradient-free ideas naturally extend to combinatorial and low-dimensional spaces, achieving strong performance at a fraction of the cost. By making fast BO available across a range of input spaces, we hope practitioners can choose BO over faster methods (e.g.\ \texttt{CMA-ES}) for a much wider class of problems, even when a single evaluation takes only a second. Historically, such gains in efficiency have repeatedly broadened what BO can do: max-value entropy search \citep{wang2017max} made information-theoretic acquisition cheap enough for routine use, GPU-based GP inference \citep{gardner2018gpytorch} made GPs practical at much larger scales, and differentiable Monte Carlo acquisition functions \citep{balandat2020botorch} made batch, multi-objective, and constrained BO tractable. We hope our fast BO can play a similar role for black-box optimization.

\section{Background}

\subsection{Bayesian Optimization}
Bayesian optimization \citep{garnett2023bayesian} is concerned with finding the global optimum $\x^\star \in \Xcal$ of a function $f:\Xcal \rightarrow \R$, where $f$ is assumed to be unknown and can only be evaluated pointwise. Standard BO literature also defines $f$ as being \say{expensive} (e.g.\ in time or money), though there is no clear cutoff between \say{inexpensive} and \say{expensive}. Moreover, $\Xcal$ can refer to both continuous and combinatorial input spaces; we specify which one is meant in each section.

To try to find the global optimum $\x^\star$, BO typically proceeds as follows: (i) approximate $f$ by fitting a (probabilistic) \emph{surrogate model} on the observed data points $\mathcal{D}_t$, and (ii) acquire the next point by solving for $\x_{t+1} = \argmax_{\x \in \Xcal} \alpha_t (\x)$. Here, $\alpha_t(\cdot)$ is called an \emph{acquisition function} and relies on the surrogate model to trade off exploration and exploitation.

\paragraph{Surrogate Models.} Although a wide range of surrogate models have been used in BO \citep{hutter2011sequential,bergstra2011algorithms}, by far the most common are still Gaussian processes \citep[GPs;][]{williams2006gaussian}. GPs are specified only by a mean function $\mu(\x)$ and a covariance function or kernel $k(\x, \x'): \Xcal \times \Xcal \rightarrow \R$. For certain kernels, GPs can also be expressed as Bayesian linear regression. The simplest example is the \emph{linear kernel}, i.e.\
\begin{equation}\label{eq:lin-kernel}
  k_\mathrm{linear}(\x, \x') := b_0 + b_1 \x^\top \x',
\end{equation}
where $b_0, b_1 > 0$ are hyperparameters.

For this kernel, the GP or equivalent Bayesian linear model has a posterior mean and variance (defined at a test feature $\uu$) given by
\begin{align}
  \label{eq:post-mean} m(\uu) &= c + \bar\w^\top\uu, &
  \bar\w &= \Z^\top \Ktil^{-1}(\y - c\one), \\
  \label{eq:post-var} s^2(\uu) &= \uu^\top\Sig\,\uu, &
  \Sig &= \I_M - \Z^\top\Ktil^{-1}\Z.
\end{align}
Here, $\uu$ denotes the features $\uu(\x)$ of size $M$, $\w\sim\mathcal N(\zero,\I_M)$ is a weight prior, $c$ is the constant mean, and $\sigma^2$ is the homoscedastic noise variance. Moreover, $\Z\in\R^{N\times M}$ represents the stacked training features and $\Ktil := \Z\Z^\top + \sigma^2\I_N$.

As can be seen from \Cref{eq:post-mean,eq:post-var}, the linear kernel has a posterior mean that is \emph{linear} in $\uu$ and a posterior variance that is a \emph{convex} quadratic (i.e.\ $\Sig\succeq 0$). These facts are exploited in \Cref{sec:acq-cont,sec:acq-comb}.

\paragraph{Acquisition Functions.} \label{para:af-intro}
Similar to surrogate models, many acquisition functions (AFs) have been used in BO over the years, with TS, UCB, and EI (defined below) arguably being the three most commonly used.

\emph{Thompson Sampling} \citep[TS;][]{thompson1933likelihood} consists of a randomized acquisition
\begin{equation} \label{eq:af-ts}
  \alpha_{\mathrm{TS}}(\x) \;=\; \tilde f_t(\x),
  \qquad \tilde f_t(\cdot) \sim p(f\mid\mathcal D_t).
\end{equation}
At each BO iteration, the goal is to optimize a draw $\tilde f_t$ from the model's posterior $p(f\mid\mathcal D_t)$ at round $t$ (i.e.\ a sample path for a GP).

\emph{Upper Confidence Bound} \citep[UCB;][]{srinivas2010gaussian} maximizes an explicit trade-off between posterior mean and posterior standard deviation, namely
\begin{equation} \label{eq:af-ucb}
  \alpha_{\mathrm{UCB}}(\x) \;=\; m(\x) \;+\; \beta\, s(\x),
\end{equation}
where $\beta>0$ is an exploration hyperparameter.

\emph{Expected Improvement} \citep[EI;][]{movckus1975bayesian,jones1998efficient} measures the expected gain over the incumbent $y_{\max}$ (the best value observed)
\begin{equation} \label{eq:af-ei}
  \alpha_{\mathrm{EI}}(\x)
  \;=\; \mathbb{E}\left[\,\big(f(\x)-y_{\max}\big)^+ \,\right],
\end{equation}
using the model's predictive distribution $f(\x)$.

\subsection{Linear Structure in BO Algorithms}

Our fast BO methods are inspired by two existing BO algorithms: Linear BO from \citet{doumont2026linear} for continuous variables, and \texttt{CoCaBO} from \citet{ru2020bayesian} for combinatorial variables.

\paragraph{Linear BO.}
\citet{doumont2026linear} show that, for high-dimensional problems, a (spherically projected) linear surrogate can achieve state-of-the-art performance, thereby questioning the need for more complicated models (e.g.\ non-linear GPs). Their surrogate is a GP with a spherical-linear (SL) kernel (equivalent to Bayesian linear regression, see previous section):
\begin{equation} \label{eq:spher-lin-kernel}
    k_\mathrm{SL}\left(\x, \x'\right) := b_0 + b_1 \; P\left(\z\right)^{\top} P\left(\z'\right),
\end{equation}
where $b_i > 0$ are hyperparameters with $b_0 + b_1 = 1$, and $P: \R^D \rightarrow \Sphere^D$ is a bijective mapping onto the unit sphere (minus its north pole). Here, $\z$ is a scaled version of $\x$ (and $\z'$ of $\x'$), i.e.\ 
\begin{equation} \label{eq:input-scaling}
    \z := \tfrac{1}{a} \begin{bmatrix} \tfrac{x_1}{\ell_1} & \cdots  & \tfrac{x_D}{\ell_D} \end{bmatrix}^\top
\end{equation}
for hyperparameters $a, \ell_1, \dots, \ell_D > 0$. The authors choose $P$ to be the inverse stereographic projection:
\begin{equation} \label{eq:stereo-proj}
    P(\z) := \tfrac{1}{\|\z\|^2 + 1} \begin{bmatrix}2 z_1 & \cdots  & 2 z_D & \|\z\|^2 - 1\end{bmatrix}^\top.
\end{equation}

\paragraph{\texttt{CoCaBO}.} \label{para:cocabo-intro}
\citet{ru2020bayesian} present a new BO approach, called \texttt{CoCaBO}, to select values for both categorical and continuous inputs. For categorical inputs, the method relies on a categorical kernel equivalent to the overlap kernel \citep{dreczkowski2023framework} or Hamming kernel \citep{kondor2002diffusion}, namely
\begin{equation} \label{eq:cocabo-kernel}
  k_\mathrm{CoCaBO}(\x,\x') := \frac{b}{D}\sum_{i=1}^{D}\mathds{1}[\,x_i = x_i'\,],
\end{equation}
with $b>0$ being a hyperparameter (the kernel variance). In \Cref{sec:model-comb}, we show how this kernel can be seen as the combinatorial counterpart to the linear kernel in \Cref{eq:lin-kernel}.

\subsection{Related Work}
Our method is related to two other works, both relying on Linear BO \citep{doumont2026linear} for faster acquisitions. The main difference between these works and ours is the search space considered: for continuous inputs, ours considers the hypercube $\Xcal := [-1,1]^D$, while theirs focuses on either the sphere $\Xcal := \Sphere^D$ or the unbounded $\Xcal := \R^D$. Both works explicitly mention our setting, but leave it for future work. Lastly, we also provide an extension of our method to combinatorial spaces $\Xcal := \prod_{i=1}^D \Xcal_i$ (with $\Xcal_i$ being discrete sets).

\paragraph{Fast Latent-Space BO.} \citet{fan2026search} explore BO for optimization of latent spaces in generative models: they focus on $\Xcal := \Sphere^D$, where high-dimensional latents concentrate. Their fitting procedure is similar to ours (though not fully, e.g.\ their lengthscale only rescales the features, and thus never requires a new eigendecomposition), but their AF optimization is fundamentally different, due to the difference in search space.

\paragraph{Acquisition Geometry of Spherical Linear BO.} \citet{whitehead2026frontier} analyzes the AF optimization of Linear BO on $\Xcal := \R^D$, where acquisitions can take place on the entire sphere (minus its north pole). While black-box problems are rarely unbounded, this result is a first step in an interesting direction: in this setting, maximizing EI reduces exactly to a one-dimensional search. This idea is similar in spirit to our MM algorithm for EI (\Cref{sec:acq-cont}), where each step solves a UCB-type problem for some weight $\beta$.

\section{Method (Continuous)} \label{sec:method-cont}
In this section, we assume $\Xcal := [-1,1]^D$, which is simply a centered version of the standard $[0,1]^D$.%

\subsection{Surrogate Model} \label{sec:model-cont}
Since a GP with a linear kernel is equivalent to Bayesian linear regression, its posterior can be computed either in $O(DN^2+N^3)$ time through a function-space view, or in $O(ND^2+D^3)$ time through a weight-space view \citep{williams2006gaussian}. This property is already exploited in \citet{doumont2026linear} and is not a new contribution of our fast BO method.

However, the main speed bottleneck of Linear BO is not its computation of the posterior, but rather the fitting of its hyperparameters. Because of its ARD lengthscales, Linear BO needs to fit $D+4$ hyperparameters: $D$ lengthscales $\ell_i$, one global lengthscale $a$, one mixture weight $b_1$, one noise variance $\sigma^2$, and one constant mean $c$. To fit all these hyperparameters, the method relies on gradient-based marginal-likelihood maximization, which can be very slow.

\paragraph{Faster Fit.} Instead, our fast BO methods aim to reduce the number of hyperparameters, after which fitting can be performed quickly with gradient-free methods. Specifically, we slightly alter the surrogate model from Linear BO by dropping ARD (and its hyperprior), leaving effectively only one lengthscale: the global lengthscale. Although this change drastically reduces the number of hyperparameters (from hundreds or thousands to only four), it has little effect on performance \citep[see Figure~8 in][]{doumont2026linear}.

To reduce fitting overhead, we want to avoid repeated eigendecompositions during marginal-likelihood optimization. Similar to \citet{fan2026search}, a single eigendecomposition can be reused when fitting hyperparameters that leave the features unchanged (e.g.\ $b_1$ or $\sigma^2$). In our setting, however, the global lengthscale $a$ is placed inside the spherical projection $P$, which means new values of $a$ require new eigendecompositions. As a solution, we propose a mixed approach: perform a grid search over $a$ (parallelizing the eigendecompositions), and solve the inner optimization of the remaining three hyperparameters. The latter can be done analytically for $c$, and using Nelder--Mead (\citeyear{nelder1965simplex}) for $b_1$ and $\sigma^2$ (while reusing existing eigendecompositions). Lastly, to further reduce the number of eigendecompositions, we run the full grid search every $10$ BO iterations; in between, we hold $a$ fixed and repeat only the inner fit. Further details are given in \Cref{app:fit,app:impl}.

\paragraph{Standardized Notation.} Finally, to simplify the task of AF optimization in the next section, we rewrite Linear BO (with or without ARD) as
\begin{equation} \label{eq:feat-kernel-cont}
  k_\mathrm{SL}(\x,\x') = \uu(\x)^\top\uu(\x'),
\end{equation}
where $b_0$, $b_1$, and $P(\z)$ have been combined into a single vector $\uu$, namely
\begin{equation} \label{eq:feat-cont}
  \uu(\x) := \big[\sqrt{b_1}\,P(\z);\ \sqrt{b_0}\,\big] \in \Sphere^{D+1}.
\end{equation}
As a result, Linear BO can now more easily be understood in terms of the properties of \Cref{eq:post-mean,eq:post-var}: its mean and variance are linear and convex quadratic in $\uu(\x)$, respectively. However, note that $\uu(\x)$ is not linear in $\x$, due to $P(\cdot)$. 

\subsection{Acquisition Functions} \label{sec:acq-cont}

Using this standardized notation for our spherical-linear surrogate, the three AFs from \Cref{para:af-intro} simplify substantially. In fact, as will be shown below, the three different optimization problems are intertwined: UCB and EI only differ by one parameter, and require repeatedly solving TS-based subproblems. For simplicity of notation, we will often consider maximizing directly for $\z \in [-1/a,1/a]^D$, since $\z = \x / a$.
\paragraph{(a) Thompson Sampling (TS).}
Under our linear surrogate from \Cref{eq:feat-kernel-cont}, the TS acquisition function in \Cref{eq:af-ts} becomes simply
\begin{equation} \label{eq:lin-max-cont}
  \alpha_{\mathrm{TS}}(\z)=c+\w^\top\uu(\z), \quad \text{with } \w\sim\mathcal N(\bar\w,\Sig).
\end{equation}
This acquisition $\alpha_{\mathrm{TS}}(\x)$ is \emph{linear} in the features $\uu(\x)$.
Through \Cref{eq:stereo-proj,eq:feat-cont}, we have
\begin{equation} \label{eq:frac-max-cont}
  \alpha_{\mathrm{TS}}(\z) = \mathrm{const} + \frac{2 \g^\top\z + \kappa (\|\z\|^2-1)}{\|\z\|^2+1},
\end{equation}
where $\g := \sqrt{b_1}\,\w_{1:D}$ and $\kappa := \sqrt{b_1}\, w_{D+1}$. Here, $\mathrm{const}$ denotes a term independent of $\z$. \Cref{eq:frac-max-cont} corresponds to a \emph{single-ratio quadratic fractional program} in $\z$, and can therefore be solved with \Ldink{\textsc{Dinkelbach's algorithm}} \citep{dinkelbach1967nonlinear}. We refer the reader to \Cref{app:deriv-ts} for an introduction to this algorithm and the derivation of \Cref{eq:frac-max-cont}, and to \citet{schaible1976fractional} or \citet{stancu1997fractional} for details.

Dinkelbach's algorithm is guaranteed to converge to the global optimum, without requiring gradients, line searches or restarts. As a result, TS is maximized exactly over the hypercube $\Xcal$ (up to numerical tolerance), whereas TS in high dimensions typically relies on approximate optimizers \citep{rashidi2024cylindrical,fan2026adaptive}. The algorithm converges superlinearly (often within a few iterations), with each iteration costing $O(D)$.

\paragraph{(b) Upper Confidence Bound (UCB).} For our linear surrogate, UCB can be written as
\begin{align}
  \alpha_{\mathrm{UCB}}(\z) &= m(\uu) \;+\; \beta\, s(\uu), \notag\\
  \label{eq:ucb-cont} &= c + \bar\w^\top\uu \;+\; \beta\, \|\Sig^{1/2}\uu\|,
\end{align}
using \Cref{eq:post-mean,eq:post-var}. This objective consists of two parts: $\bar\w^\top\uu$ is linear in $\uu$ (and can thus be solved with Dinkelbach's algorithm), while $\|\Sig^{1/2}\uu\|$ is convex in $\uu$. Since UCB needs to be maximized and contains a convex component, we can use the classic \Lmm{\textsc{Minorize--Maximize (MM)}} approach \citep{hunter2004tutorial} to solve this optimization problem.

\begin{algorithm}[t]
\caption{Acquisition optimization (continuous). For our experiments: $K=T=10$.}
\label{alg:acq-mini}
\begin{algorithmic}[1]
\Statex \textbf{(a) TS}
\vspace{.25\baselineskip}
\State $\w\sim\mathcal N(\bar\w,\Sig)$
\vspace{.25\baselineskip}
\State \Return \Ldink{$\textsc{Dinkelbach}(\w)$}
\vspace{-.45\baselineskip}
\Statex \hrulefill
\Statex \makebox[5.3em][l]{\textbf{(b) UCB:}} $\lambda(\uu)=\beta$
\Statex \makebox[5.3em][l]{\textbf{(c) EI:}} $\lambda(\uu)=\varphi(\gamma)/\Phi(\gamma)$, \ $\gamma=\frac{m(\uu)-y_{\max}}{s(\uu)}$
\vspace{.25\baselineskip}
\State draw $\x_1,\dots,\x_K\sim\mathcal U(\Xcal)$
\vspace{.25\baselineskip}
\For{$j=1,\dots,T$}\Comment{\Lmm{MM iterations}}
  \For{$k=1,\dots,K$ \textbf{in parallel}}\Comment{\Lmm{restarts}}
    \State $\uu_k\gets\uu(\x_k)$
    \State \Lmm{$\w_k\gets\bar\w+\tfrac{\lambda(\uu_k)}{s(\uu_k)}\Sig\uu_k$}
    \State $\x_k\gets\Ldink{\textsc{Dinkelbach}(\w_k)}$
  \EndFor
\EndFor
\vspace{.25\baselineskip}
\State \Return $\argmax\alpha(\x_k)$
\end{algorithmic}
\end{algorithm}

To find a solution $\z^\star$, the MM procedure iteratively updates $\z$ as follows:
\begin{equation} \label{eq:MM-argmax}
   \z_{j+1} \in \argmax_{\z\in[-1/a,1/a]^D} h(\z\mid\z_j),
\end{equation}
where $h(\z\mid\z_j)$ is called the minorizer (or majorizer for minimization problems). This minorizer has to satisfy two conditions:
\begin{align*}
  h(\z\mid\z_0) &\le \alpha_{\mathrm{UCB}}(\z) \text{ for all } \z, \tag*{(global minorant)}\\
  h(\z_0\mid\z_0) &= \alpha_{\mathrm{UCB}}(\z_0) \tag*{(tightness)}.
\end{align*}
Moreover, in the MM algorithm, the minorizer has to be solved repeatedly. For a fast algorithm, $h(\z\mid\z_j)$ should therefore be cheap to maximize. We refer the reader to \citet{lange2016mm} for a more in-depth treatment of MM methods.

To construct a function $h(\z\mid\z_j)$ that meets these criteria, we transform $\alpha_{\mathrm{UCB}}(\z)$ into a linear function of $\uu(\z)$. This linear function can then quickly be solved for $\z$ using Dinkelbach's algorithm (as done for TS), allowing us to efficiently compute $\z_{j+1}$ in \Cref{eq:MM-argmax}.
 
To begin, we look at the first-order tangent of the seminorm $s(\uu)=\|\Sig^{1/2}\uu\|$. For $s(\uu_0)>0$, convexity gives
\begin{align*}
  s(\uu) &\ge s(\uu_0) + \nabla s(\uu_0)^\top (\uu - \uu_0) \\
  &= s(\uu_0) + \tfrac{(\Sig\uu_0)^\top}{s(\uu_0)} (\uu - \uu_0).
\end{align*}
We then define our minorizer $h(\z\mid\z_0)$ as \Cref{eq:ucb-cont}, except we replace $s(\uu)$ by its tangent, i.e.\
\begin{equation*}
  h(\z\mid\z_0) := c + \bar\w^\top\uu(\z) + \beta\Big[s(\uu_0) +
  \tfrac{(\Sig\uu_0)^\top(\uu(\z)-\uu_0)}{s(\uu_0)}\Big].
\end{equation*}
After simplification (see \Cref{app:mm}), we can rewrite this equation as
\begin{equation} \label{eq:mm-minorizer-cont}
  h(\z\mid\z_0) = c + \w_\mathrm{eff}(\uu_0)^\top \uu(\z),
\end{equation}
with
\begin{equation} \label{eq:mm-weight-cont}
  \w_\mathrm{eff}(\uu_0)
  := \bar\w + \frac{\lambda}{s(\uu_0)}\,\Sig\uu_0 .
\end{equation}
\Cref{eq:mm-minorizer-cont} has exactly the same structure as \Cref{eq:lin-max-cont}, and can thus be solved efficiently using Dinkelbach's algorithm. For UCB, $\lambda = \beta$.

A further rearrangement (see \Cref{app:mm}) gives
\begin{equation} \label{eq:mm-offset-cont}
  h(\z\mid\z_0) \;=\; \alpha_{\mathrm{UCB}}(\z_0) + \w_\mathrm{eff}(\uu_0)^\top
  \big(\uu(\z)-\uu_0\big),
\end{equation}
which makes the tightness condition explicit, while the global minorant condition follows from the convexity of $s$. The resulting MM algorithm guarantees nondecreasing acquisition values and stationary limit points (see \Cref{app:mm}). In practice, we use multiple random starts to seek better solutions, since the procedure does not guarantee a global maximum. See \Cref{app:impl} for implementation details.
 
\paragraph{(c) Expected Improvement (EI).}
Since our linear surrogate has a Gaussian posterior, EI can be written in closed form \citep{jones1998efficient}: 
\begin{equation}\label{eq:ei-cont}
  \alpha_{\mathrm{EI}}(\z) = \left[m(\uu) - y_{\max}\right] \Phi(\gamma) + s(\uu) \varphi(\gamma),
\end{equation}
with $\gamma := [m(\uu) - y_{\max}] / s(\uu)$. As is common, $\Phi$ and $\varphi$ represent the standard normal cumulative distribution function (CDF) and density.

Using \Cref{eq:ei-cont}, it can be shown (\Cref{app:deriv-ei}) that EI is convex and increasing in $m(\uu)$ and $s(\uu)$. Therefore, the MM procedure can once again be used to solve this optimization problem. As for UCB, we use first-order tangents to construct a minorizer $h(\z\mid\z_0)$ that satisfies both criteria.

Again, we can write $h(\z\mid\z_0)$ as a linear function of $\uu$ (see \Cref{app:deriv-ei}):
\begin{equation} \label{eq:ei-minorizer-cont}
  h(\z\mid\z_0) = \Phi(\gamma_0)\,\big[\,c - y_{\max} + \w_\mathrm{eff}(\uu_0)^\top\uu(\z)\big].
\end{equation}
Here, $\w_\mathrm{eff}$ is defined as in \Cref{eq:mm-weight-cont}, except we now have $\lambda = \varphi(\gamma_0)/\Phi(\gamma_0)$. Since $\Phi(\gamma_0) > 0$, maximizing \Cref{eq:ei-minorizer-cont} is equivalent to solving \Cref{eq:lin-max-cont}, and can thus be solved efficiently using Dinkelbach's algorithm.

In fact, EI and UCB can be solved using the same algorithm (and implementation), with only one line changed: instead of keeping the standard UCB exploration hyperparameter $\beta$ (e.g.\ $\beta=2$ in our experiments), $\lambda$ needs to be recomputed at every iteration in terms of $\gamma_0$. This also shows how EI acts as UCB with a self-tuning exploration weight ($\lambda$ changes over time instead of being fixed at the beginning). Our three AFs are summarized in \Cref{alg:acq-mini}.

\section{Method (Combinatorial)} \label{sec:method-comb}
For this section, similar to \citet{doumont2025omnipresent}, we define the search space $\Xcal$ as a product $\Xcal := \prod_{i=1}^D \Xcal_i$ of finite sets $\Xcal_i$. Here, each set $\Xcal_i$ denotes a categorical variable with $|\Xcal_i|$ categories.
 
\subsection{Surrogate Model} \label{sec:model-comb}

As noted in \Cref{para:cocabo-intro}, the \texttt{CoCaBO} kernel in \Cref{eq:cocabo-kernel} can be seen as the combinatorial counterpart to our linear kernel from \Cref{sec:model-cont} (but without the spherical projection). In fact, as we show below, it can also be written as an inner product.

\paragraph{Standardized Notation.} First, we define $\vphi(x_i)\in\{0,1\}^{|\Xcal_i|}$ as the one-hot encoding of categorical variable $x_i$ over its category set $\Xcal_i$. Then, we concatenate the different $\vphi(x_i)$ into $\vphi(\x)\in\{0,1\}^{M}$, with $M=\sum_{i=1}^{D}|\Xcal_i|$. Using $\vphi(\x)$, we obtain
\begin{equation} \label{eq:feat-kernel-comb}
  k_\mathrm{CoCaBO}(\x,\x') = \uu(\x)^\top\uu(\x'),
\end{equation}
where
\begin{equation} \label{eq:feat-comb}
  \uu(\x) := \sqrt{b/D}\;\vphi(\x).
\end{equation}
As a result, \Cref{eq:cocabo-kernel} can be written in a similar way to \Cref{eq:feat-kernel-cont}, allowing us to reuse the results from \Cref{eq:post-mean,eq:post-var}: the posterior mean and variance are again linear and convex quadratic in $\uu$, respectively. As before, we will use these facts for fast AF optimization in \Cref{sec:acq-comb}.

Nevertheless, we highlight one major difference between $k_\mathrm{CoCaBO}$ (combinatorial inputs) and $k_\mathrm{SL}$ (continuous inputs): $k_\mathrm{CoCaBO}$ does not need to map the inputs $\x$ to the unit sphere $\Sphere^{D+1}$. For combinatorial inputs, acquisition optimization only takes place over a discrete set $\Xcal$, meaning the boundary-seeking behavior from \citet{doumont2026linear} no longer applies (see \Cref{app:deriv-comb-norm}). Lastly, there is no constant mean $c$ and no rescaled variable $\z$ here, since $\x$ is discrete.

\paragraph{Faster Fit.} By rewriting $k_\mathrm{CoCaBO}$ as an inner product, it follows that the resulting GP is actually a (Bayesian) linear regression. Therefore, the posterior can also be computed in $O(NM^2+M^3)$ time through a weight-space view, which is useful when $N > M$. To the best of our knowledge, we are the first to make this connection evident for $k_\mathrm{CoCaBO}$.

Since $k_\mathrm{CoCaBO}$ has no lengthscale inside $\uu(\x)$, fitting its hyperparameters is faster and simpler than for $k_\mathrm{SL}$. Its two hyperparameters, the kernel variance $b$ and the noise variance $\sigma^2$, can be fitted with MacKay's fixed-point updates \citep{mackay1992bayesian}, reusing a single eigendecomposition of $\mathbf{H}^\top\mathbf{H}$, where $\mathbf{H}$ stacks the one-hot encodings of the training inputs (see \Cref{app:fit}).

\subsection{Acquisition Functions} \label{sec:acq-comb}

Using our standardized notation, we can solve the three AFs in almost the same way as for continuous inputs: UCB and EI can be solved using the MM approach, relying on repeated solves of TS-based subproblems. Again, both AFs only differ in $\lambda$. The main difference with the continuous setting is that TS and related subproblems do not require Dinkelbach's algorithm (there is no spherical projection), and can instead be solved through a simple per-variable argmax.

\paragraph{(a) Thompson Sampling (TS).}
For the \texttt{CoCaBO} kernel, the TS acquisition function from \Cref{eq:af-ts} becomes simply
\begin{equation} \label{eq:ts-comb}
  \alpha_{\mathrm{TS}}(\x)
  = \w^\top\uu(\x),
  \quad \text{with } \w\sim\mathcal N(\bar\w,\Sig).
\end{equation}
Since each $\vphi(x_i)$ is a one-hot encoding, the acquisition simplifies to an additive function:
\begin{equation} \label{eq:ts-separable-comb}
  \alpha_{\mathrm{TS}}(\x)
  = \sqrt{b/D}\;\sum_{i=1}^{D}\w_i^\top\vphi(x_i).
\end{equation}
Here, $\alpha_{\mathrm{TS}}(\x)$ is separable, so its global maximizer is
  simply the per-variable (discrete) argmax, which can be found in $O(M)$ time. In our experiments, all three AFs are optimized within the Hamming trust region of \texttt{MCBO} \citep{dreczkowski2023framework}, where each subproblem is solved exactly (see \Cref{app:deriv-ts,app:impl}).

\paragraph{(b) UCB and (c) EI.}
As for continuous inputs, UCB and EI are both convex and increasing in $m(\uu)$ and $s(\uu)$, so we can again rely on the MM algorithm. For UCB ($\lambda = \beta$), this gives the minorizer
\begin{align}
  \label{eq:mm-minorizer-comb} h(\x\mid\x_0) &= \w_\mathrm{eff}(\uu_0)^\top\uu(\x) + \mathrm{const}, \\
  \text{with}\quad \w_\mathrm{eff}(\uu_0)  &= \bar\w + \frac{\lambda}{s(\uu_0)}\,\Sig\,\uu_0, \notag
\end{align}
which is identical to \Cref{eq:mm-minorizer-cont,eq:mm-weight-cont}, except that $\x$ replaces $\z$ and the constant term is different (see \Cref{app:mm}). For EI, the minorizer is the same up to a positive factor $\Phi(\gamma_0)$, with $\lambda = \varphi(\gamma_0)/\Phi(\gamma_0)$ recomputed at every MM iteration, as in the continuous case (see \Cref{app:deriv-ei}). For both AFs, maximizing $h(\x\mid\x_0)$ is again equivalent to TS in \Cref{eq:ts-comb}, so its global maximum can be found through a per-variable argmax (i.e.\ without Dinkelbach's algorithm).

\section{Experiments} \label{sec:experiments}

\paragraph{Setup.} For high-dimensional continuous inputs, we follow the setup of \citet{doumont2026linear}: six benchmarks from $60$D to $6392$D, $30$ initial Sobol points, and a budget of $1000$ evaluations (with batch size $q=1$). To highlight the breadth of our proposed algorithms, we add four low-dimensional benchmarks ($12$D to $33$D, $500$ evaluations) and five combinatorial benchmarks ($25$ to $125$ variables, $200$ evaluations). All continuous benchmarks are served through \texttt{Bencher} \citep{papenmeier2025bencher}, whereas the combinatorial ones are taken from \texttt{MCBO} \citep{dreczkowski2023framework}. For both continuous and combinatorial inputs, we compare against state-of-the-art BO methods --- including Linear BO and \texttt{CoCaBO} themselves --- as well as other well-known black-box optimizers (e.g.\ evolutionary and genetic algorithms). Random search is included throughout. Further details on the benchmarks and baselines are given in \Cref{app:benchmarks,app:baselines}.

For all these problems, evaluating a single point typically takes less than a second. While some problems are simple proxies for slower real-world processes (e.g.\ the \texttt{MuJoCo} and \texttt{Robot}~\texttt{Pushing} simulators), this is not true for all of them. For example, hyperparameter tuning on \texttt{Lasso-DNA} and \texttt{SVM}, or the combinatorial \texttt{LABS} and \texttt{MaxSAT} problems, all describe actual tasks of interest, highlighting the need for black-box optimizers covering all areas of the Pareto frontier.
 
\begin{figure*}[t]
  \centering
  \setlength{\abovecaptionskip}{3pt}
  \includegraphics[scale=0.806]{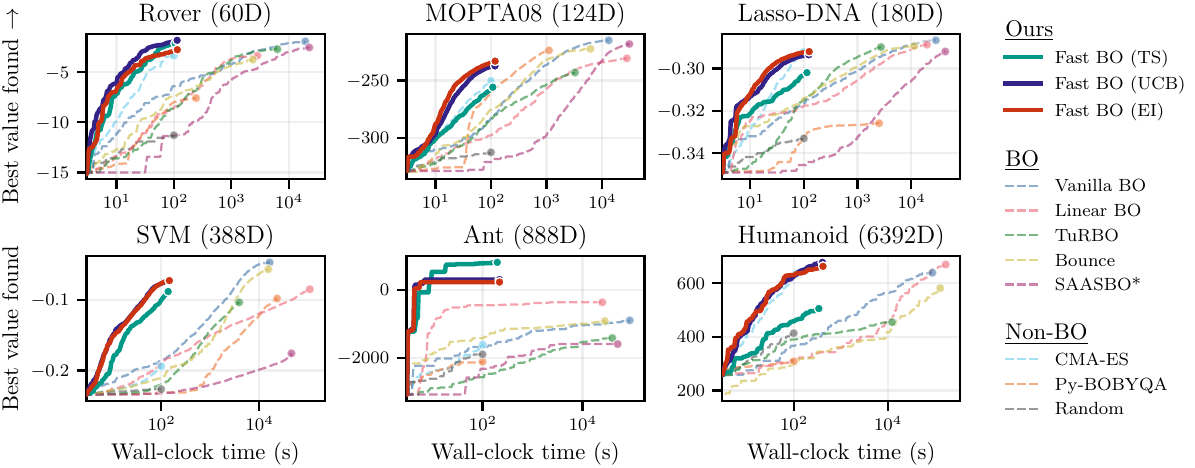}
  \caption{Best value found against wall-clock time on the high-dimensional benchmarks, with each method run for 1000 evaluations (except \texttt{SAASBO}$^{*}$ for 500). For this figure, evaluations are assumed to take only $0.1$ seconds. Even in this fast regime, \textbf{the speed advantage of \texttt{CMA-ES} does not lead to better results compared to our fast BO methods}: its overall wall-clock time is similar to ours, while performing worse (on average).}
  \label{fig:anytime}
\end{figure*}

\paragraph{Metrics.} On Pareto plots, we report the final best value against the total \emph{algorithm time}, i.e.\ the total wall-clock time of the algorithm, excluding black-box evaluations. All methods run on CPU, except \texttt{SAASBO}$^{*}$, which always runs on GPU, and some baselines on \texttt{Humanoid} (see \Cref{app:hardware}). Both the final best value and the algorithm time are averaged over $10$ seeds for high-dimensional inputs, $40$ seeds for low-dimensional inputs, and $20$ seeds for combinatorial inputs. \Cref{fig:pareto-aggregate} summarizes each setting in a single panel, with per-benchmark results in \Cref{fig:pareto-cont-high,fig:pareto-cont-low,fig:pareto-comb}. Besides Pareto plots, we also provide plots of the best value per iteration in \Cref{app:sota}, and total wall-clock time (including evaluations of $0.1$ seconds) in \Cref{fig:anytime}.

\subsection{Continuous Problems} \label{sec:exp-cont}

\paragraph{High Dimensions ($D > 50$).}
On the summary plot in \Cref{fig:pareto-aggregate}, our fast UCB and EI methods achieve performance similar to the strongest method, Vanilla BO, while being about $700\times$ faster (on average). We ablate the components behind this speedup in \Cref{app:ablation}. For EI, this strong performance is not surprising: our surrogate and AF from \Cref{sec:method-cont} are very close to those of Linear BO, which \citet{doumont2026linear} have already reported to match Vanilla BO. Moreover, as concluded by the authors in their Figure~9: \say{replacing EI with UCB does not significantly affect performance}, which matches our findings for Fast BO. Their Figure~10 also shows TS trailing EI, as it does for our methods. Furthermore, our results extend the conclusion made by \citet{doumont2026linear}: high-dimensional BO methods are matched or outperformed by a linear model \emph{with only four hyperparameters} (instead of hundreds or thousands).

Viewing BO through the lens of a speed-performance frontier reveals another interesting result for these high-dimensional problems: \texttt{CMA-ES} achieves a similar sample efficiency (on average) to \texttt{TuRBO}, in contrast to its poor performance reported by \citet{hvarfner2024vanilla}. However, its speed advantage only matters for evaluations in the milliseconds: already at $0.1$ seconds per evaluation, its total wall-clock time is similar to ours, while its final values are worse (see \Cref{fig:anytime}).

\paragraph{Low/Mid Dimensions ($D < 50$).} To adapt our methods to lower-dimensional spaces, we make two key modifications: (i) we only fit our model on the last $D$ observations (to prevent saturation of our linear model), and (ii) we adopt the trust region from \texttt{TuRBO} \citep{eriksson2019scalable}. Since the latter requires ARD lengthscales to set the side lengths of the trust region, we fit Linear BO \citep{doumont2026linear} every $50$ iterations, and use its lengthscales for the trust region (and for nothing else). By doing so, we achieve competitive performance while being about $10\times$ faster than \texttt{TuRBO} and about $100\times$ faster than Vanilla BO (see \Cref{fig:pareto-aggregate}): our TS method performs on par with Vanilla BO and \texttt{Bounce}, and only \texttt{TuRBO} performs slightly better. We provide an ablation of these two added components in \Cref{app:ablation-lowdim}. Interestingly, for these low-dimensional problems, TS is now better than EI and UCB, instead of worse. We hypothesize that this is due to \texttt{TuRBO}'s trust region, which might favor more explorative behavior (\texttt{TuRBO} itself relies on TS). Lastly, inside a small trust region, our linear model acts as a local-linear approximation, hinting at a connection with local BO \citep{muller2021local,nguyen2022local} and model-based trust-region methods such as \texttt{BOBYQA} \citep{powell2009bobyqa}.

\subsection{Combinatorial Problems} For a fair speed comparison, we run all \texttt{MCBO} \citep{dreczkowski2023framework} methods through a faster, batched \texttt{PyTorch} \citep{paszke2019pytorch} implementation instead of the original \texttt{pandas} \citep{mckinney2010data} one, which gives the exact same results (see \Cref{app:baselines}). Our UCB and EI methods reach performance close to the state-of-the-art One-hot BO \citep{doumont2025omnipresent}, i.e.\ a GP with an RBF kernel on one-hot inputs, combined with the genetic algorithm of \texttt{MCBO}, while being one to two orders of magnitude faster. Although our fast methods rely on the \texttt{CoCaBO} kernel, they perform significantly better than the original method. This difference comes from AF optimization, which \citet{dreczkowski2023framework} showed to be a key driver of performance: when we optimize our AF with \texttt{CoCaBO}'s bandit instead, our performance drops to that of \texttt{CoCaBO} (see \Cref{app:ablation}).

\section{Conclusion}
By recognizing and subsequently exploiting linear structures in well-known BO methods, we address one of the main remaining challenges of BO: its computational overhead.  Through novel applications of classic linear algebra and optimization techniques, we develop a standardized framework that speeds up both model fitting and AF optimization. On high- and low-dimensional problems across continuous and combinatorial inputs, our fast BO methods consistently cover new areas of the Pareto frontier, achieving strong performance while being faster by several orders of magnitude. As a result, BO can now be applied to evaluations as cheap as $0.1$ seconds, while still outperforming faster black-box optimizers such as \texttt{CMA-ES}. More broadly, we hope our fast methods will make BO the default choice for a much wider range of black-box problems.

Moreover, besides these practical gains, our methods also open up new avenues for theoretical analysis. For example, TS is now maximized exactly (as assumed by its regret bounds; \citealp{russo2014learning}), and EI can be studied as UCB with a self-tuning exploration weight. Our trust-region variant also suggests links to local BO and model-based trust-region methods, and our exact solvers serve as a reference for studying inexact AF optimization in gradient-based BO.

\paragraph{Limitations.} Our experiments do not cover very low-dimensional spaces ($D < 10$), mixed spaces, or batch acquisitions, and extending our methods to these settings, as well as to other AFs, is left for future work. Lastly, for expensive evaluations, our speed advantage no longer matters, and practitioners may prefer the slightly more consistent performance of slower, state-of-the-art BO methods (e.g.\ Vanilla BO on high-dimensional problems; see \Cref{fig:pareto-cont-high}).

\newpage
\bibliography{references}

\clearpage
\appendix
\crefalias{section}{appendix}
\crefalias{subsection}{subappendix}
\crefalias{subsubsection}{subsubappendix}
\thispagestyle{empty}
\onecolumn
\aistatstitle{Appendix}

\section{Derivations} \label{app:derivations}

\subsection{Surrogate Fits} \label{app:fit}

\paragraph{Continuous Inputs.}
We standardize the targets and set $b_0 = 1 - b_1$, as in \Cref{eq:spher-lin-kernel}. This leaves four hyperparameters: the global lengthscale $a$, the mixture weight $b_1$, the noise variance $\sigma^2$, and the constant mean $c$. We fit these hyperparameters by minimizing the classic negative log marginal likelihood (NLL)
\begin{equation*}
  \mathrm{NLL}(a, b_1, \sigma^2, c) = \frac12 (\y - c\one)^\top \Ktil^{-1} (\y - c\one) + \frac12 \log\det\Ktil + \frac N2 \log 2\pi.
\end{equation*}
Using \Cref{eq:feat-cont}, the Gram matrix $\Ktil$ from \Cref{eq:post-mean,eq:post-var} can be written as
\begin{equation*}
  \Ktil = b_1\PP\PP^\top + b_0\one\one^\top + \sigma^2\I_N,
\end{equation*}
where $\PP \in \R^{N\times(D+1)}$ stacks the projected training inputs $P(\z_1),\dots,P(\z_N)$. Computed naively, every evaluation of the NLL costs $O(N^3)$, due to the inverse and the determinant of $\Ktil$. Since an optimizer evaluates the NLL many times, we rewrite it below such that each evaluation only costs $O(\min(N,D))$. Note that $a$ is the only hyperparameter inside $\PP$. Therefore, for a fixed value of $a$, we only require a single eigendecomposition. This is similar to \citet{fan2026search}, who fit a linear model on the sphere with a single eigendecomposition. Unlike their model, however, ours has a lengthscale inside the projection and a constant feature $b_0$, which is why we need the grid search over $a$ and the rank-one correction below.

To begin, we split $\Ktil$ into two parts, i.e.\
\begin{equation*}
  \Ktil = \A + b_0\one\one^\top, \qquad \text{with} \quad \A := b_1\PP\PP^\top + \sigma^2\I_N.
\end{equation*}
For a fixed $a$, we compute the eigendecomposition $\PP\PP^\top = Q\,\mathrm{diag}(\nu_1,\dots,\nu_N)\,Q^\top$ once. In this eigenbasis, $\A$ becomes diagonal for any $b_1$ and $\sigma^2$:
\begin{equation*}
  \A^{-1} = Q\Del Q^\top, \quad \text{with} \quad \Del := \mathrm{diag}\big((b_1\nu_i + \sigma^2)^{-1}\big)_{i=1}^N, \qquad \text{and} \qquad \log\det\A = \sum_{i=1}^N \log(b_1\nu_i + \sigma^2).
\end{equation*}

Next, the remaining part $b_0\one\one^\top$ only has rank one. Using the Sherman--Morrison formula and the matrix determinant lemma, we get
\begin{equation*}
  \Ktil^{-1} = \A^{-1} - \frac{b_0}{1 + b_0 \eta_{11}}\,\A^{-1}\one\one^\top\A^{-1}, \qquad \log\det\Ktil = \log\det\A + \log(1 + b_0 \eta_{11}),
\end{equation*}
with the scalar $\eta_{11} := \one^\top\A^{-1}\one$. Similarly, we define the two scalars $\eta_{1y} := \one^\top\A^{-1}\y$ and $\eta_{yy} := \y^\top\A^{-1}\y$.

Lastly, the NLL can be minimized analytically for the constant mean $c$. Setting its derivative with respect to $c$ to zero gives the generalized least-squares estimate
\begin{equation*}
  c = \frac{\one^\top\Ktil^{-1}\y}{\one^\top\Ktil^{-1}\one} = \frac{\eta_{1y}}{\eta_{11}}.
\end{equation*}
For this value of $c$, we have $\one^\top\A^{-1}(\y - c\one) = \eta_{1y} - c\eta_{11} = 0$. The rank-one term therefore drops out of the quadratic form, which simplifies to $\eta_{yy} - \eta_{1y}^2/\eta_{11}$. Putting everything together, the NLL becomes (up to a constant)
\begin{equation} \label{eq:fit-nll}
  \mathrm{NLL}(b_1,\sigma^2) = \frac12\Big(\eta_{yy} - \frac{\eta_{1y}^2}{\eta_{11}}\Big)
  + \frac12\sum_{i=1}^N\log(b_1\nu_i + \sigma^2)
  + \frac12\log(1 + b_0\,\eta_{11}).
\end{equation}
Here, the three scalars $\eta_{11}$, $\eta_{1y}$ and $\eta_{yy}$ only require the cached vectors $\hat\one := Q^\top\one$ and $\hat\y := Q^\top\y$, since
\begin{equation*}
  \eta_{11} = \hat\one^\top\Del\hat\one, \qquad \eta_{1y} = \hat\one^\top\Del\hat\y, \qquad \eta_{yy} = \hat\y^\top\Del\hat\y.
\end{equation*}
As $\Del$ is diagonal, each evaluation of \Cref{eq:fit-nll} thus costs only $O(N)$, without any matrix operations. We minimize it over $(b_1,\sigma^2)$ with Nelder--Mead, and choose $a$ with a grid search around this inner fit (see \Cref{alg:fit}, and \Cref{app:impl} for the settings).

So far, we have used the eigendecomposition of the $N\times N$ matrix $\PP\PP^\top$, whose cost grows cubically with the number of observations. When $N > D+1$, we can instead work with the smaller $(D+1)\times(D+1)$ matrix $\PP^\top\PP$, which has the same nonzero eigenvalues (with eigenvectors $\PP\vv_i/\sqrt{\nu_i}$, where $\vv_i$ are the eigenvectors of $\PP^\top\PP$). The remaining $N-D-1$ eigenvalues of $\PP\PP^\top$ are all zero. Ordering the eigenvalues such that the first $D+1$ are nonzero, $\eta_{11}$ can then be split as
\begin{equation*}
  \eta_{11} = \sum_{i=1}^{N} \frac{\hat 1_i^2}{b_1\nu_i + \sigma^2}
  = \sum_{i=1}^{D+1} \frac{\hat 1_i^2}{b_1\nu_i + \sigma^2} + \frac{1}{\sigma^2}\sum_{i=D+2}^{N} \hat 1_i^2.
\end{equation*}
Since $Q$ is orthogonal, we have $\|\hat\one\|^2 = \|\one\|^2 = N$, such that the last sum equals $N - \sum_{i=1}^{D+1}\hat 1_i^2$. As a result, $\eta_{11}$ only requires the $D+1$ nonzero eigenvectors, and the same holds for $\eta_{1y}$ and $\eta_{yy}$ (using $\one^\top\y$ and $\y^\top\y$). Overall, the eigendecomposition costs $O(ND\min(N,D) + \min(N,D)^3)$ for each value of $a$, after which each NLL evaluation costs only $O(\min(N,D))$.

\paragraph{Combinatorial Inputs.}
With the features from \Cref{eq:feat-comb}, the model is a Bayesian linear regression
\begin{equation*}
  y = \vphi(\x)^\top\w + \epsilon,
  \qquad \w \sim \mathcal N(\zero, \tau^{-1}\I_M),
  \qquad \epsilon \sim \mathcal N(0, \sigma^2),
\end{equation*}
with prior precision $\tau := D/b$ (i.e.\ we absorb the scale $\sqrt{b/D}$ of \Cref{eq:feat-comb} into the weights, which is the usual form of MacKay's updates). As before, the targets are standardized, but the model has no constant mean. This leaves two hyperparameters: $\tau$ and $\sigma^2$. Let $\mathbf{H} \in \{0,1\}^{N\times M}$ stack the one-hot encodings of the training inputs. In weight space, the NLL can then be written as \citep{bishop2006pattern}
\begin{equation*}
  \mathrm{NLL}(\tau, \sigma^2) = \frac{1}{2\sigma^2}\|\y - \mathbf{H}\bar\w\|^2 + \frac\tau2\|\bar\w\|^2
  + \frac12\log\det\big(\tau\I_M + \sigma^{-2}\mathbf{H}^\top\mathbf{H}\big)
  - \frac M2\log\tau + \frac N2\log\sigma^2 + \frac N2\log 2\pi,
\end{equation*}
where $\bar\w = \sigma^{-2}(\tau\I_M + \sigma^{-2}\mathbf{H}^\top\mathbf{H})^{-1}\mathbf{H}^\top\y$ is the posterior mean of $\w$. This is the same NLL as for continuous inputs (with $c = 0$), except written in weight space, where MacKay's updates are easiest to derive (see below).

Unlike for continuous inputs, the features $\vphi(\x)$ do not depend on any hyperparameter. Therefore, similar to \citet{fan2026search}, both hyperparameters can be fitted using a single eigendecomposition $\mathbf{H}^\top\mathbf{H} = V\,\mathrm{diag}(\omega_1,\dots,\omega_M)\,V^\top$. In this eigenbasis, the matrix $\tau\I_M + \sigma^{-2}\mathbf{H}^\top\mathbf{H}$ is diagonal for any $\tau$ and $\sigma^2$, with entries $\tau + \omega_i/\sigma^2$. As a result, its log determinant, $\bar\w$ and the residual $\|\y - \mathbf{H}\bar\w\|^2$ all cost only $O(M)$, using the cached vector $V^\top\mathbf{H}^\top\y$ and the scalar $\y^\top\y$.

Lastly, instead of a generic optimizer, we minimize the NLL with MacKay's fixed-point updates \citep{mackay1992bayesian,bishop2006pattern}. Setting the derivatives of the NLL with respect to $\tau$ and $\sigma^2$ to zero, and rearranging, gives
\begin{equation*}
  \tau = \frac{N_\mathrm{eff}}{\|\bar\w\|^2},
  \qquad
  \sigma^2 = \frac{\|\y - \mathbf{H}\bar\w\|^2}{N - N_\mathrm{eff}},
  \qquad \text{with} \quad
  N_\mathrm{eff} := \sum_{i=1}^M \frac{\omega_i}{\sigma^2\tau + \omega_i}.
\end{equation*}
Here, $N_\mathrm{eff}$ can be seen as the effective number of parameters. Since $\bar\w$ and $N_\mathrm{eff}$ depend on $\tau$ and $\sigma^2$ themselves, we iterate these two equations as updates until they converge. Each update reuses the same eigendecomposition, and costs only $O(M)$.

\paragraph{No Projection for Combinatorial Inputs.} \label{app:deriv-comb-norm}
For continuous inputs, the spherical projection gives every input the same prior variance. Since $\|P(\z)\| = 1$, we have
\begin{equation*}
  k_\mathrm{SL}(\x,\x) = \|\uu(\x)\|^2 = b_0 + b_1 = 1
  \quad \text{for all } \x.
\end{equation*}
Without the projection, the prior variance of a linear kernel grows with $\|\x\|$, which pushes the acquisitions towards the boundary of the search space \citep{doumont2026linear}. For combinatorial inputs, however, the one-hot features already have a constant norm. Each $\vphi(x_i)$ has exactly one nonzero entry, so $\|\vphi(\x)\|^2 = D$, and therefore
\begin{equation*}
  k_\mathrm{CoCaBO}(\x,\x) = \|\uu(\x)\|^2 = b
  \quad \text{for all } \x\in\Xcal.
\end{equation*}
As a result, no projection is needed.

\subsection{Thompson Sampling} \label{app:deriv-ts}

TS consists of two steps: (i) drawing a posterior sample $\w \sim \mathcal N(\bar\w, \Sig)$, and (ii) maximizing the resulting sample path $\w^\top\uu(\z)$ over the search box. As we show below, both steps reuse the fit from \Cref{app:fit}, and the second step can be solved exactly.

\paragraph{Posterior Samples.}
To draw $\w$, we use Matheron's rule \citep{wilson2020efficiently}, i.e.\
\begin{equation*}
  \w = \w_0 + \Z^\top\Ktil^{-1}\big(\y - c\one - \Z\w_0 - \bm\epsilon\big),
  \qquad \text{with} \quad \w_0 \sim \mathcal N(\zero, \I_M) \quad \text{and} \quad \bm\epsilon \sim \mathcal N(\zero, \sigma^2\I_N),
\end{equation*}
where $M = D+2$ (see \Cref{eq:feat-cont}). This gives an exact posterior sample, by correcting the prior sample $\w_0$ with the posterior update. Moreover, $\Ktil^{-1}$ can be applied using the eigendecomposition and the rank-one correction from \Cref{app:fit}, such that $\Sig$ never has to be formed.

\paragraph{Fractional Program.}
Using \Cref{eq:stereo-proj,eq:feat-cont}, the sample path can be written out as
\begin{equation*}
  \w^\top\uu(\z) = \sqrt{b_1}\,\w_{1:D}^\top\frac{2\z}{\|\z\|^2+1}
  + \sqrt{b_1}\,w_{D+1}\frac{\|\z\|^2-1}{\|\z\|^2+1}
  + \sqrt{b_0}\,w_{D+2}.
\end{equation*}
Here, the last term does not depend on $\z$. With $\g = \sqrt{b_1}\,\w_{1:D}$ and $\kappa = \sqrt{b_1}\,w_{D+1}$, the first two terms can be combined into the single ratio
\begin{equation} \label{eq:ts-ratio}
  R(\z) := \frac{A(\z)}{B(\z)},
  \qquad \text{with} \quad
  A(\z) := 2\g^\top\z + \kappa\big(\|\z\|^2-1\big)
  \quad \text{and} \quad
  B(\z) := \|\z\|^2+1,
\end{equation}
which gives \Cref{eq:frac-max-cont}. As a result, maximizing TS is equivalent to maximizing $R(\z)$ over the search box $\mathcal Z := [\z_\mathrm{lo},\z_\mathrm{hi}]$ in scaled coordinates (i.e.\ $[-1/a,1/a]^D$ in \Cref{sec:acq-cont}). This corresponds to a single-ratio fractional program, which can be solved with Dinkelbach's algorithm (see below).

\paragraph{Dinkelbach's Algorithm.}
Consider a general problem of the form $\max_{\z\in\mathcal Z} A(\z)/B(\z)$, where $A$ and $B$ are continuous, $\mathcal Z$ is compact, and $B > 0$ on $\mathcal Z$. Instead of maximizing this ratio directly, Dinkelbach's algorithm relies on the parametric function
\begin{equation*}
  F(\rho) := \max_{\z\in\mathcal Z}\; A(\z) - \rho\, B(\z),
\end{equation*}
which is continuous and strictly decreasing in $\rho$. If $\rho^\star$ denotes the optimal ratio, we have $A(\z) - \rho^\star B(\z) \le 0$ for all $\z \in \mathcal Z$ (since $B > 0$), with equality at the maximizers of the ratio. As a result, $\rho^\star$ is the unique root of $F$, and both problems have the same maximizers at $\rho = \rho^\star$.

To find this root, Dinkelbach's algorithm starts from a feasible point $\z_0$, with $\rho_0 = A(\z_0)/B(\z_0)$, and then alternates between the two steps
\begin{equation*}
  \z_n \in \argmax_{\z\in\mathcal Z}\; A(\z) - \rho_n B(\z),
  \qquad
  \rho_{n+1} = \frac{A(\z_n)}{B(\z_n)}.
\end{equation*}
Here, the ratios $\rho_n$ are nondecreasing: since $\rho_n$ is the ratio of a feasible point, we have $F(\rho_n) \ge 0$, and thus $A(\z_n) - \rho_n B(\z_n) \ge 0$, i.e.\ $\rho_{n+1} \ge \rho_n$. Moreover, the ratios converge superlinearly to $\rho^\star$, as long as each subproblem in the first step is solved globally \citep{dinkelbach1967nonlinear,schaible1976fractional}. In our case, we start from $\z_0 = \zero$, i.e.\ the center of the box (or of the trust region).

\paragraph{Exact Subproblems.}
For $R(\z)$, we can group the terms in $\|\z\|^2$ as
\begin{equation*}
  \kappa\big(\|\z\|^2-1\big) - \rho\big(\|\z\|^2+1\big) = (\kappa - \rho)\|\z\|^2 - (\kappa + \rho),
\end{equation*}
which gives the subproblem
\begin{equation*}
  A(\z) - \rho\, B(\z) = \sum_{i=1}^D \Big[\, 2g_iz_i - (\rho - \kappa)\, z_i^2 \,\Big] - (\kappa + \rho).
\end{equation*}
As can be seen, this subproblem is separable over the coordinates. If $\rho > \kappa$, each term is a concave quadratic in $z_i$, with maximizer
\begin{equation*}
  z_i = \mathrm{clip}\Big(\frac{g_i}{\rho - \kappa},\; z_{\mathrm{lo},i},\; z_{\mathrm{hi},i}\Big).
\end{equation*}
Otherwise, each term is convex (or linear), and its maximum lies at one of the two endpoints $z_{\mathrm{lo},i}$ and $z_{\mathrm{hi},i}$. As a result, each subproblem can be solved exactly in $O(D)$ time, and Dinkelbach's algorithm returns the global maximizer of TS over the box.

\paragraph{Combinatorial Inputs.}
For combinatorial inputs, the posterior covariance of $\w$ is diagonal in the eigenbasis of the fit (see \Cref{app:fit}). A posterior sample can thus be drawn as
\begin{equation*}
  \w = \bar\w + V\,\mathrm{diag}\Big(\frac{1}{\sqrt{\tau + \omega_1/\sigma^2}},\dots,\frac{1}{\sqrt{\tau + \omega_M/\sigma^2}}\Big)\,\bm\xi, \qquad \text{with} \quad \bm\xi \sim \mathcal N(\zero, \I_M).
\end{equation*}
Since \Cref{eq:ts-separable-comb} is a sum of per-variable terms, its global maximizer sets each $x_i$ to the category with the largest entry in $\w_i$, which takes $O(M)$ time. Within a Hamming trust region of radius $r$ around a center $\bar\x$ (see \Cref{app:impl}), let $\delta_i \ge 0$ be the gain of this best category over $\bar x_i$. Since the objective is separable and the constraint only counts the number of changed variables, the constrained maximizer changes the (up to) $r$ variables with the largest positive gains (i.e.\ replacing a chosen variable by one with a larger gain can only improve the objective). These variables can be found by sorting, in $O(M + D\log D)$ time.

\subsection{Upper Confidence Bound} \label{app:mm}

In this section, we derive the minorizer of UCB from \Cref{sec:acq-cont}, show that it satisfies both MM conditions, and discuss the convergence of the resulting MM algorithm.

\paragraph{Minorizer.}
Since $\sigma^2 > 0$, the posterior variance is strictly positive, i.e.\ $s(\uu_0) > 0$, and the tangent of $s$ at $\uu_0$ is always defined:
\begin{equation*}
  s(\uu) \ge s(\uu_0) + \frac{(\Sig\uu_0)^\top(\uu - \uu_0)}{s(\uu_0)}.
\end{equation*}
Replacing $s(\uu)$ in \Cref{eq:ucb-cont} by this tangent, and collecting the terms in $\uu(\z)$, we get
\begin{equation*}
  h(\z\mid\z_0)
  = c + \w_\mathrm{eff}(\uu_0)^\top\uu(\z)
  + \beta\Big[s(\uu_0) - \frac{\uu_0^\top\Sig\uu_0}{s(\uu_0)}\Big].
\end{equation*}
Here, the term in brackets vanishes, since $\uu_0^\top\Sig\uu_0 = s(\uu_0)^2$, which gives \Cref{eq:mm-minorizer-cont,eq:mm-weight-cont}. Using the same identity, we also have
\begin{equation*}
  c + \w_\mathrm{eff}(\uu_0)^\top\uu_0
  = c + \bar\w^\top\uu_0 + \beta\, s(\uu_0)
  = \alpha_\mathrm{UCB}(\z_0),
\end{equation*}
such that adding and subtracting $\w_\mathrm{eff}(\uu_0)^\top\uu_0$ in \Cref{eq:mm-minorizer-cont} gives \Cref{eq:mm-offset-cont}.

\paragraph{MM Conditions.}
The minorizer $h(\z\mid\z_0)$ satisfies both MM conditions from \Cref{sec:acq-cont}. First, since $s$ is convex in $\uu$, its tangent lies below $s$ everywhere in feature space, such that $h(\z\mid\z_0) \le \alpha_\mathrm{UCB}(\z)$ for all $\z$ (global minorant). Second, \Cref{eq:mm-offset-cont} gives $h(\z_0\mid\z_0) = \alpha_\mathrm{UCB}(\z_0)$ (tightness). Here, convexity is only required in $\uu$, and not in $\z$.

\paragraph{Convergence.}
Combining both MM conditions gives
\begin{equation*}
  \alpha_\mathrm{UCB}(\z_{j+1})
  \;\ge\; h(\z_{j+1}\mid\z_j)
  \;\ge\; h(\z_j\mid\z_j)
  \;=\; \alpha_\mathrm{UCB}(\z_j),
\end{equation*}
where the second inequality holds because $\z_{j+1}$ maximizes $h(\cdot\mid\z_j)$. As a result, the acquisition values are nondecreasing, and converge since they are bounded on the compact box. Moreover, since $s > 0$ on the box, $h$ is continuous and matches the gradient of $\alpha_\mathrm{UCB}$ at $\z_0$. These are the conditions of \citet[Theorem~1]{razaviyayn2013unified}, which guarantee that every accumulation point of the iterates is a stationary point of UCB over the box. Since this is only a local guarantee, we run MM from multiple random starts (see \Cref{app:impl}). Each MM step costs one Dinkelbach solve, as for TS (see \Cref{app:deriv-ts}), plus $O(ND)$ to compute $\Sig\uu_0$.

\paragraph{Combinatorial Inputs.}
For combinatorial inputs, we follow the same practice as \texttt{MCBO} \citep{dreczkowski2023framework}, and define $s$ as the predictive standard deviation
\begin{equation*}
  s(\uu) = \sqrt{\uu^\top\Sig\uu + \sigma^2},
\end{equation*}
i.e.\ the norm of $[\Sig^{1/2}\uu;\,\sigma]$, which is still convex in $\uu$ and has the same gradient $\Sig\uu/s(\uu)$. Repeating the steps above without the constant mean $c$, the term in brackets no longer vanishes, since $\uu_0^\top\Sig\uu_0 = s(\uu_0)^2 - \sigma^2$. Instead, we obtain \Cref{eq:mm-minorizer-comb} with
\begin{equation*}
  \mathrm{const} = \beta\,\frac{\sigma^2}{s(\uu_0)}.
\end{equation*}
Since this constant does not change the maximizer, each MM step is again equivalent to TS (see \Cref{app:deriv-ts}), also within the Hamming trust region. Lastly, since $\Sig \succ 0$ and $\sigma^2 > 0$, $s$ is strictly convex, such that every MM step that changes the configuration strictly increases the acquisition. Since $\Xcal$ is finite, MM therefore stops at a fixed point after a finite number of steps. Here, computing $\Sig\uu_0$ costs $O(M^2)$ per step.

\subsection{Expected Improvement} \label{app:deriv-ei}

Similar to UCB, we first show that EI is convex and increasing in $m$ and $s$, which then allows us to derive a minorizer that is linear in $\uu$.

\paragraph{Convexity and Monotonicity.}
For $\xi \sim \mathcal N(0,1)$, EI can be written as
\begin{equation*}
  \mathrm{EI}(m,s) = \mathbb E\big[(m - y_{\max} + s\,\xi)^+\big].
\end{equation*}
Since the function inside the expectation is convex in $(m,s)$ for each fixed $\xi$, EI is also convex in $(m,s)$. Moreover, for $s > 0$, its partial derivatives are
\begin{equation*}
  \frac{\partial\,\mathrm{EI}}{\partial m} = \Phi(\gamma) > 0,
  \qquad
  \frac{\partial\,\mathrm{EI}}{\partial s} = \varphi(\gamma) > 0,
\end{equation*}
such that EI is increasing in both $m$ and $s$.

\paragraph{Minorizer.}
Since EI is convex in $(m,s)$, its tangent plane at $(m_0,s_0)$ lies below it, i.e.\
\begin{equation*}
  \mathrm{EI}(m,s) \ge \mathrm{EI}(m_0,s_0) + \Phi(\gamma_0)\,(m - m_0) + \varphi(\gamma_0)\,(s - s_0).
\end{equation*}
Moreover, since $\varphi(\gamma_0) > 0$, we can further replace $s - s_0$ by its tangent lower bound from \Cref{app:mm}, i.e.\
\begin{equation*}
  s - s_0 \ge \frac{(\Sig\uu_0)^\top(\uu - \uu_0)}{s_0}.
\end{equation*}
After substituting $m(\uu) = c + \bar\w^\top\uu$ and defining $\lambda_0 := \varphi(\gamma_0)/\Phi(\gamma_0)$, collecting the terms gives the minorizer
\begin{equation*}
  h(\z\mid\z_0) = \mathrm{EI}(m_0,s_0)
  + \Phi(\gamma_0)\,\w_\mathrm{eff}(\uu_0)^\top\big(\uu(\z) - \uu_0\big),
\end{equation*}
with $\w_\mathrm{eff}$ as in \Cref{eq:mm-weight-cont}, except with $\lambda = \lambda_0$. As for UCB, this minorizer is a global minorant, and is tight at $\z_0$.

Lastly, to obtain \Cref{eq:ei-minorizer-cont}, we use the two identities
\begin{equation*}
  \mathrm{EI}(m_0,s_0) = \Phi(\gamma_0)\big[m_0 - y_{\max} + \lambda_0 s_0\big],
  \qquad
  \w_\mathrm{eff}(\uu_0)^\top\uu_0 = \bar\w^\top\uu_0 + \lambda_0 s_0.
\end{equation*}
After substituting both into $h(\z\mid\z_0)$, all terms in $\uu_0$ cancel except $c - y_{\max}$, which gives \Cref{eq:ei-minorizer-cont}. The convergence argument from \Cref{app:mm} then applies unchanged, with $\lambda = \lambda_0$ at every step.

\paragraph{Combinatorial Inputs.}
Since the posterior of $k_\mathrm{CoCaBO}$ is again Gaussian, EI has the same closed form, and the derivation above applies with $\x$ instead of $\z$. With the predictive $s$ and no constant mean, the same cancellation now leaves
\begin{equation*}
  \w_\mathrm{eff}(\uu_0)^\top\uu_0 = \bar\w^\top\uu_0 + \lambda_0\,\frac{s_0^2 - \sigma^2}{s_0},
\end{equation*}
which gives the minorizer
\begin{equation*}
  h(\x\mid\x_0) = \Phi(\gamma_0)\big[\w_\mathrm{eff}(\uu_0)^\top\uu(\x) + \mathrm{const}\big],
  \qquad \text{with } \mathrm{const} = \lambda_0\,\frac{\sigma^2}{s(\uu_0)} - y_{\max}.
\end{equation*}
Here, $\w_\mathrm{eff}$ is defined as in \Cref{eq:mm-minorizer-comb}, with $\lambda = \lambda_0$. Since $\Phi(\gamma_0) > 0$, maximizing $h(\x\mid\x_0)$ is again equivalent to TS, and is solved by a per-variable argmax (see \Cref{app:deriv-ts}).

\section{Experimental Details} \label{app:exp}

\subsection{Benchmarks} \label{app:benchmarks}

\Cref{tab:benchmarks} lists all benchmarks. All continuous benchmarks are served by \texttt{Bencher} \citep{papenmeier2025bencher}. The six high-dimensional benchmarks are those of \citet{doumont2026linear}, and the five combinatorial tasks come from \texttt{MCBO} \citep{dreczkowski2023framework}. The implementations of \texttt{LABS}, \texttt{MaxSAT} and \texttt{Contamination} follow \texttt{Bounce} \citep{papenmeier2023bounce}. \texttt{Bencher}, \texttt{MCBO}, \texttt{Bounce} and \texttt{BoTorch} are all released under the MIT License, as is \texttt{Gym} \citep{brockman2016openai}. \texttt{MuJoCo} is released under the Apache License~2.0, \texttt{pycma} under the BSD 3-Clause License, and \texttt{Py-BOBYQA} under the GNU GPL~v3. Continuous runs start from a $30$-point scrambled Sobol design \citep{sobol1967distribution}, and combinatorial runs from $20$ random points. The combinatorial tasks are minimized, and all their variables are binary, except for \texttt{Pest}, whose $25$ variables have $5$ categories each.

Every evaluation is seeded from the seed of the run, so all runs are reproducible. \texttt{Rover}, \texttt{Robot}~\texttt{Pushing} and the \texttt{MuJoCo} \citep{todorov2012mujoco} tasks are stochastic, i.e.\ repeated evaluations of the same point give different values. For the \texttt{MuJoCo} tasks, each evaluation averages $10$ rollouts, as in \citet{wang2020learning}. For \texttt{Lunar}~\texttt{Lander}, each evaluation averages $50$ rollouts with fixed seeds, shared by all points, as in \citet{eriksson2019scalable}. Repeated evaluations of the same point therefore give the same value. These rollout counts differ from the defaults of \texttt{Bencher} (one episode for \texttt{Lunar}~\texttt{Lander}, one rollout for \texttt{Ant}, and five for the other \texttt{MuJoCo} tasks). We instead follow the settings described in the original papers \citep{wang2020learning,eriksson2019scalable}.

\begin{table*}[t]
\centering\footnotesize
\caption{Benchmarks, with their dimension $D$, their number of evaluations (including the initial design), and the number of rollouts averaged per evaluation (-- for deterministic benchmarks).}
\label{tab:benchmarks}
\vspace{2pt}
\begin{tabular}{@{}l l r r r l@{}}
\toprule
& Benchmark & $D$ & Evaluations & Rollouts & Reference \\
\midrule
\multirow{6}{*}{\shortstack[l]{High-dimensional\\continuous}}
& \texttt{Rover}     & $60$   & $1000$ & $1$  & \citet{wang2018batched,eriksson2019scalable} \\
& \texttt{MOPTA08}   & $124$  & $1000$ & --   & \citet{eriksson2021high} \\
& \texttt{Lasso-DNA} & $180$  & $1000$ & --   & \citet{nardi2022lassobench} \\
& \texttt{SVM}       & $388$  & $1000$ & --   & \citet{eriksson2021high} \\
& \texttt{Ant}       & $888$  & $1000$ & $10$ & \citet{wang2020learning} \\
& \texttt{Humanoid}  & $6392$ & $1000$ & $10$ & \citet{wang2020learning} \\
\midrule
\multirow{4}{*}{\shortstack[l]{Low-dimensional\\continuous}}
& \texttt{Lunar}~\texttt{Lander}   & $12$ & $500$ & $50$ & \citet{brockman2016openai,eriksson2019scalable} \\
& \texttt{Robot}~\texttt{Pushing}  & $14$ & $500$ & $1$  & \citet{wang2018batched,eriksson2019scalable} \\
& \texttt{Swimmer}                 & $16$ & $500$ & $10$ & \citet{wang2020learning} \\
& \texttt{Hopper}                  & $33$ & $500$ & $10$ & \citet{wang2020learning} \\
\midrule
\multirow{5}{*}{Combinatorial}
& \texttt{Contamination} & $25$  & $200$ & -- & \citet{baptista2018bayesian} \\
& \texttt{Pest}          & $25$  & $200$ & -- & \citet{oh2019combinatorial} \\
& \texttt{LABS}          & $50$  & $200$ & -- & \citet{packebusch2016low} \\
& \texttt{MaxSAT60}      & $60$  & $200$ & -- & \citet{oh2019combinatorial} \\
& \texttt{MaxSAT125}     & $125$ & $200$ & -- & \citet{oh2019combinatorial} \\
\bottomrule
\end{tabular}
\end{table*}

\subsection{Baselines} \label{app:baselines}

\paragraph{Continuous.}
We compare against random search, Vanilla BO \citep{hvarfner2024vanilla}, \texttt{TuRBO} \citep{eriksson2019scalable}, \texttt{SAASBO} \citep{eriksson2021high}, \texttt{Bounce} \citep{papenmeier2023bounce} and Linear BO \citep{doumont2026linear}, which are all implemented with \texttt{BoTorch} \citep{balandat2020botorch}. We also compare against two derivative-free methods, \texttt{CMA-ES} \citep{hansen2001completely}, using the \texttt{pycma} implementation \citep{hansen2019pycma}, and \texttt{Py-BOBYQA} \citep{cartis2019improving}, an implementation of \texttt{BOBYQA} \citep{powell2009bobyqa}.

Here, \texttt{SAASBO}$^{*}$ denotes the \texttt{BoTorch}~0.18 implementation of \texttt{SAASBO}, which uses a new and faster JAX-based sampler for its fully Bayesian inference. Since this inference remains very expensive, \texttt{SAASBO}$^{*}$ always runs on a GPU, with a budget of only $500$ evaluations. For the high-dimensional benchmarks (budget $1000$), its final value and total algorithm time are therefore not comparable to those of the other methods. As a result, we omit \texttt{SAASBO}$^{*}$ from the high-dimensional Pareto plots (\Cref{fig:pareto-aggregate,fig:pareto-cont-high}) and the time breakdown (\Cref{fig:time-breakdown}), and only show it in the figures that compare methods along the run (\Cref{fig:sota,fig:time-per-step,fig:anytime}). In these figures, it has no runs on \texttt{Humanoid}, since its memory requirements exceed our $40$\,GB GPUs. On the low-dimensional benchmarks (budget $500$), \texttt{SAASBO}$^{*}$ runs as many iterations as the other baselines, and is therefore included in both the Pareto and per-iteration figures (\Cref{fig:pareto-aggregate,fig:pareto-cont-low,fig:sota-lowdim}).

Similarly, \texttt{TuRBO}, \texttt{Bounce} and Linear BO run on a GPU for the \texttt{Humanoid} problem. Moreover, \texttt{Bounce} starts from its own $5$-point design, as in its original paper \citep{papenmeier2023bounce}. Lastly, \texttt{Py-BOBYQA} interpolates a quadratic through $2D{+}1$ points, which exceeds the evaluation budget on \texttt{Ant} and \texttt{Humanoid}. On these two benchmarks, the entire run thus consists of its initial design, and its low algorithm time only reflects that it never starts optimizing.

\paragraph{Combinatorial.}
Our main BO baseline is One-hot BO \citep{doumont2025omnipresent}, i.e.\ a GP with an RBF kernel on one-hot inputs, for which we use its implementation on top of \texttt{MCBO} (with EI and the genetic algorithm of \texttt{MCBO} as AF optimizer). Moreover, we compare against \texttt{CoCaBO} \citep{ru2020bayesian} with the default settings of \texttt{MCBO}, i.e.\ a GP with the overlap kernel and EI, where the categories are chosen by a multi-armed bandit. Lastly, the model-free baselines consist of random search and the four optimizers of \texttt{MCBO} (hill climbing, simulated annealing, a genetic algorithm, and a multi-armed bandit), all with their default settings.

For a fair speed comparison, we run all \texttt{MCBO} methods through a faster, batched \texttt{PyTorch} implementation of the framework code, instead of the original \texttt{pandas} one. This implementation gives exactly the same results. For One-hot BO, we also report the timings obtained with the original framework code (i.e.\ \texttt{pandas} instead of \texttt{PyTorch}).

\subsection{Implementation} \label{app:impl}

All three AFs share the same fit, and the model and AF computations run in double precision. Below, we list the settings that affect our results, and refer to our code for further numerical details. We chose these settings in preliminary runs on our benchmarks, and use them unchanged on all benchmarks of a suite.

\paragraph{Fit.}
As in \Cref{sec:method-cont}, inputs are rescaled to $\Xcal = [-1,1]^D$, such that $\z = \x/a$. The global lengthscale $a$ is chosen from a grid of $10$ values
\begin{equation*}
  a_j = \sqrt{f_j D}, \qquad j = 1,\dots,10,
\end{equation*}
where the fractions $f_j$ are spaced geometrically between $10^{-3}$ and $1$. For $f_j = 1$, $a_j$ is the half-diagonal of $\Xcal$. For each value $a_j$, the inner fit from \Cref{app:fit} runs Nelder--Mead on $(b_1, \sigma^2)$, starting at $b_1 = \tfrac12$ and $\sigma^2 = 0.1$, with at most $80$ evaluations and $\sigma^2 \ge 10^{-8}$. The full grid search runs every $10$ fits. In between, $a$ is held fixed, and only the inner fit is redone (see \Cref{alg:fit}).

For combinatorial inputs, the fixed-point updates from \Cref{app:fit} run for at most $100$ iterations, also with $\sigma^2 \ge 10^{-8}$.

\begin{algorithm}[!ht]
\caption{Fit for continuous inputs. The grid search runs every tenth fit. The fits in between keep $a^\star$ fixed, and only redo the inner fit.}
\label{alg:fit}
\begin{algorithmic}[1]
\Require data $\{(\x_i,y_i)\}_{i=1}^N$; grid $\{a_j\}_{j=1}^{10}$
\For{each $a_j$ \textbf{in parallel}}\Comment{independent eigendecompositions}
  \State form $\PP(a_j)$;\quad eigendecompose the smaller of $\PP\PP^\top$ and $\PP^\top\PP$ (\Cref{app:fit})
  \State fit $(b_1,\sigma^2)$ by Nelder--Mead on \Cref{eq:fit-nll};\quad $c\gets \eta_{1y}/\eta_{11}$
\EndFor
\State $a^\star \gets \argmin_j \mathrm{NLL}(a_j)$
\State \Return posterior $(\bar\w,\Sig,c)$ at $a^\star$ via \Cref{eq:post-mean,eq:post-var}, in the same eigenbasis
\end{algorithmic}
\end{algorithm}

\paragraph{Acquisition (Continuous).}
Dinkelbach's algorithm starts at $\z = \zero$, and stops once the ratio changes by less than $10^{-6}$, or after $100$ iterations (whichever is satisfied first). For TS, we return the result of one exact solve per sample, without restarts. UCB ($\beta = 2$) and EI share the same procedure. We start from $K = 10$ points drawn uniformly in the search box, and update them together for at most $T = 10$ MM iterations. The MM iterations stop once every step is shorter than $10^{-6}$.

For EI, we use the best standardized observation as $y_{\max}$. Lastly, UCB and EI return the highest-scoring final point that has not been observed yet, or the highest-scoring one if all of them have been observed.

\paragraph{Acquisition (Combinatorial).}
All three AFs are optimized within the Hamming trust region of \texttt{MCBO} \citep{dreczkowski2023framework}. This trust region is a ball of radius $r$ around the incumbent, whose radius is adapted during the run by the trust-region manager of \texttt{MCBO}. TS returns one exact solve (see \Cref{app:deriv-ts}). UCB ($\beta = 2$) and EI follow the continuous procedure with $K = 10$ starts, where each start sets $r$ randomly chosen variables of the incumbent to random categories. These starts are then updated together for at most $10$ MM iterations, and stop once no start changes. Both AFs use the predictive standard deviation. As in the continuous case, UCB and EI return the highest-scoring final point that has not been observed yet.

Since combinatorial spaces are finite, the selected point (i.e.\ the maximizer for TS, or the best final point for UCB and EI) may already have been observed. For UCB and EI, this only happens if the final points of all $K = 10$ starts have been observed. Since our objectives are deterministic, evaluating the same point again would waste an evaluation. In that case, we instead select a nearby unobserved point, in three stages. First, we consider all unobserved points that differ from the selected point in a single variable, and choose the one with the highest acquisition value (or sampled value for TS). If all of these have been observed too, we draw up to $64$ unobserved random points that differ from the selected point in $2$ to $r$ variables, and choose the best one in the same way. Lastly, if no such point is found either, we draw random points around the center of the trust region instead. In all three stages, we only consider points inside the trust region.

\paragraph{Low-Dimensional Variant.} \label{app:tr}
On the low-dimensional benchmarks, we use the trust region of \texttt{TuRBO} \citep{eriksson2019scalable} with its default settings, and restart from a new $30$-point Sobol design when it collapses. As in \texttt{TuRBO}, its side lengths are scaled by the ARD lengthscales, normalized to a geometric mean of one. We refit these lengthscales every $50$ fits and after each restart. Moreover, as described in \Cref{sec:exp-cont}, our model is only fitted on the last $D$ observations (the window). Lastly, the spherical projection is centered at the midpoint of the trust region (i.e.\ the center of the search space considered by \texttt{TuRBO}), instead of the center of $\Xcal$.

\subsection{Hardware, Timing, and Seeds} \label{app:hardware}\label{app:stats}

\paragraph{Hardware and Timing.}
We define the algorithm time as the wall-clock time a method spends fitting its model and optimizing its AF, summed over a run. It excludes the objective evaluations and, on the continuous benchmarks, the initial fit. We use it for the $x$ axis of the Pareto figures and for \Cref{tab:timing-full}.

All CPU runs of the continuous methods used AMD EPYC 7513 CPUs with four threads, pinned to their own four cores. The combinatorial methods ran on one core of the same CPU type. For each figure, our three AFs were run with the same code and protocol, such that their timings can be compared directly. As mentioned in \Cref{app:baselines}, the GPU runs used NVIDIA A100 PCIe GPUs with $40$\,GB of memory.

\paragraph{Seeds and Statistics.}
We use $10$ seeds for the high-dimensional benchmarks, $40$ for the low-dimensional ones, and $20$ for the combinatorial ones, so as to obtain reliable means for each setting. Due to its high cost, \texttt{SAASBO}$^{*}$ only has $10$ seeds on the low-dimensional benchmarks. For a given seed, all continuous methods except \texttt{Bounce} start from the same initial design (see \Cref{app:baselines}).

In the Pareto figures, each point shows the mean final best value and the mean algorithm time over all seeds. The Pareto frontier (in gray) indicates the best mean value that can be reached within each time budget. In \Cref{fig:pareto-aggregate}, the error bars show $95\%$ intervals from a bootstrap over the seeds of each benchmark ($2000$ resamples). Unless stated otherwise, the other figures show the mean, with error bars of $\pm1$ standard error over seeds (on both axes in the per-benchmark Pareto figures). Note that many error bars are smaller than their markers, and are therefore not visible.

\section{Additional Results} \label{app:results}

\subsection{Per-Benchmark Pareto Plots} \label{app:pareto}

\begin{figure*}[t]
  \centering
  \setlength{\abovecaptionskip}{3pt}
  \includegraphics[scale=0.806]{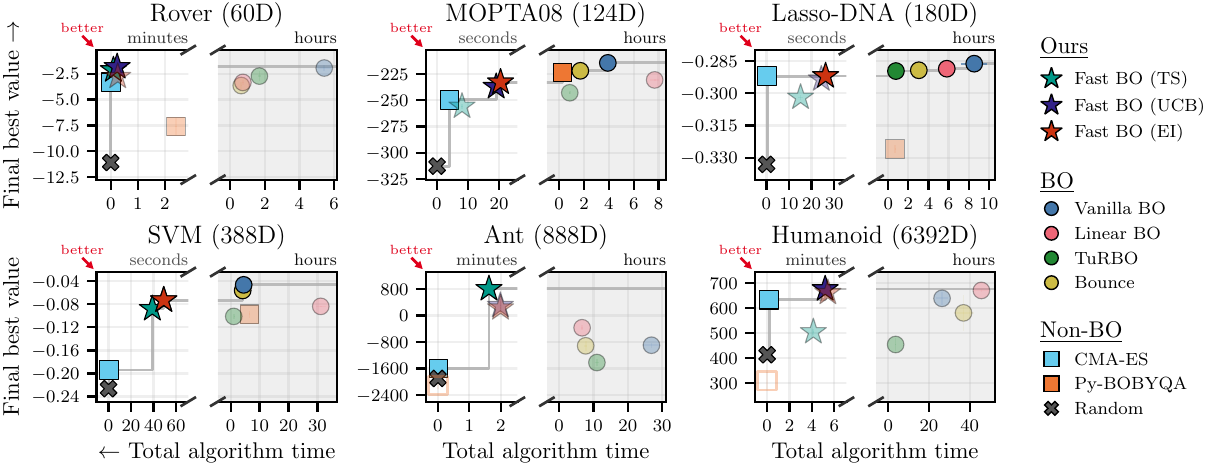}
  \caption{Extension of \Cref{fig:pareto-aggregate} for each high-dimensional benchmark. Our AFs are one to nearly four orders of magnitude faster than every model-based baseline, and always lie on the Pareto frontier. (The slow part of the time axis is shaded and uses a different scale.)}
  \label{fig:pareto-cont-high}
\end{figure*}

\begin{figure*}[t]
  \centering
  \setlength{\abovecaptionskip}{3pt}
  \includegraphics[scale=0.806]{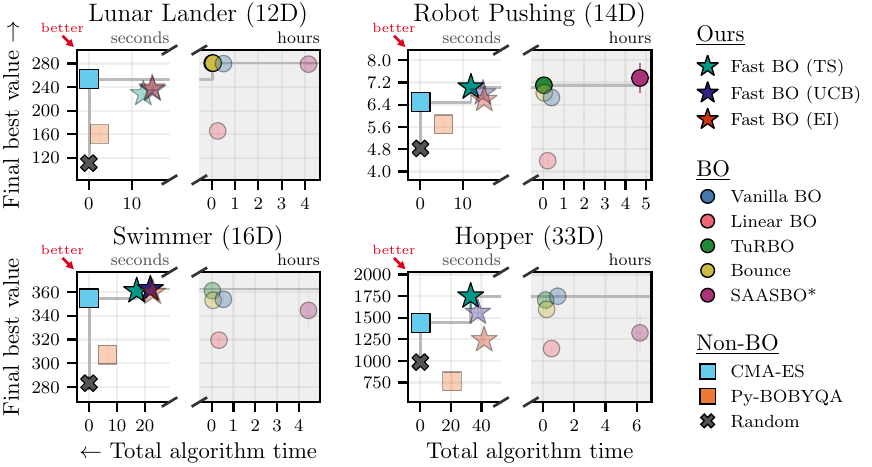}
  \caption{Extension of \Cref{fig:pareto-aggregate} for each low-dimensional benchmark. Our TS method matches or outperforms Vanilla BO on three of the four problems, at more than $100\times$ less algorithm time. (The slow part of the time axis is shaded and uses a different scale.)}
  \label{fig:pareto-cont-low}
\end{figure*}

\begin{figure*}[t]
  \centering
  \setlength{\abovecaptionskip}{3pt}
  \includegraphics[scale=0.806]{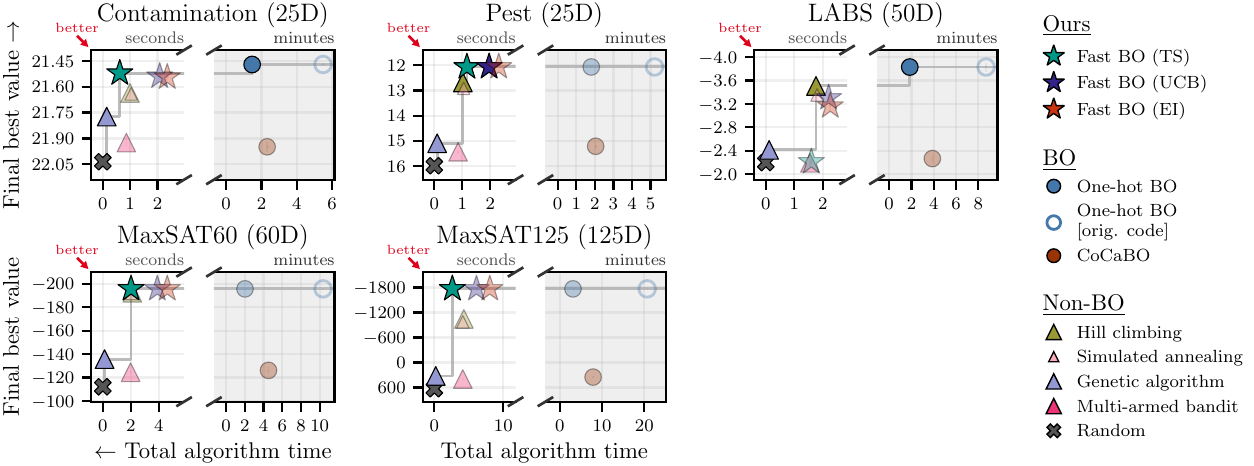}
  \caption{Extension of \Cref{fig:pareto-aggregate} for each combinatorial benchmark. Our AFs lie on the Pareto frontier on all problems except \texttt{LABS}, and are $100$--$500\times$ faster than One-hot BO with the original \texttt{MCBO} code (hollow circles). (The slow part of the time axis is shaded and uses a different scale, and the $y$ axis is inverted since the tasks are minimized.)}
  \label{fig:pareto-comb}
\end{figure*}

\Cref{fig:pareto-cont-high,fig:pareto-comb,fig:pareto-cont-low} show the Pareto plot of each benchmark, for the high-dimensional, combinatorial and low-dimensional problems, respectively. These three plots are summarized in \Cref{fig:pareto-aggregate}.

\paragraph{High-Dimensional Problems.} In \Cref{fig:pareto-cont-high}, our EI and UCB methods consistently lie on the Pareto frontier, which only holds for one other method, \texttt{CMA-ES} (besides random search). TS is faster, but usually worse than EI and UCB, as also reported by \citet{doumont2026linear}. For $N > D$, EI and UCB match or trail the state of the art, whereas for $N \approx D$ and $N < D$, they outperform all other methods (especially on \texttt{Ant}). We hypothesize that this is because, at $N < D$, the data cannot even determine a linear model, such that the extra flexibility of a GP brings little benefit.

\paragraph{Low-Dimensional Problems.} In \Cref{fig:pareto-cont-low}, our TS method lies on the Pareto frontier for three of the four problems, and is outperformed by \texttt{CMA-ES} on \texttt{Lunar}~\texttt{Lander}. On \texttt{Robot}~\texttt{Pushing}, \texttt{Swimmer} and \texttt{Hopper}, TS matches or outperforms Vanilla BO at more than $100\times$ less algorithm time. On average, UCB and EI perform slightly worse than TS (e.g.\ on \texttt{Robot}~\texttt{Pushing} and \texttt{Hopper}), likely because the trust region favors more explorative AFs (see \Cref{sec:exp-cont}).

\paragraph{Combinatorial Problems.} In \Cref{fig:pareto-comb}, our three AFs lie on the Pareto frontier on four of the five problems. On \texttt{LABS}, EI and UCB perform slightly worse than hill climbing (which lies on the Pareto frontier), and TS performs surprisingly poorly (similar to random search). We hypothesize that this is because the \texttt{LABS} objective is invariant to flipping all variables, which leaves our additive model with little signal, such that each TS sample is mostly noise. Lastly, our methods are about $20$--$100\times$ faster than One-hot BO when both use our \texttt{PyTorch} implementation of the framework code, and $100$--$500\times$ faster than One-hot BO with the original \texttt{MCBO} code (hollow circles in \Cref{fig:pareto-comb}).

\subsection{Best Value Found per Iteration} \label{app:sota}

\begin{figure*}[p]
  \centering
  \setlength{\abovecaptionskip}{3pt}
  \includegraphics[scale=0.806]{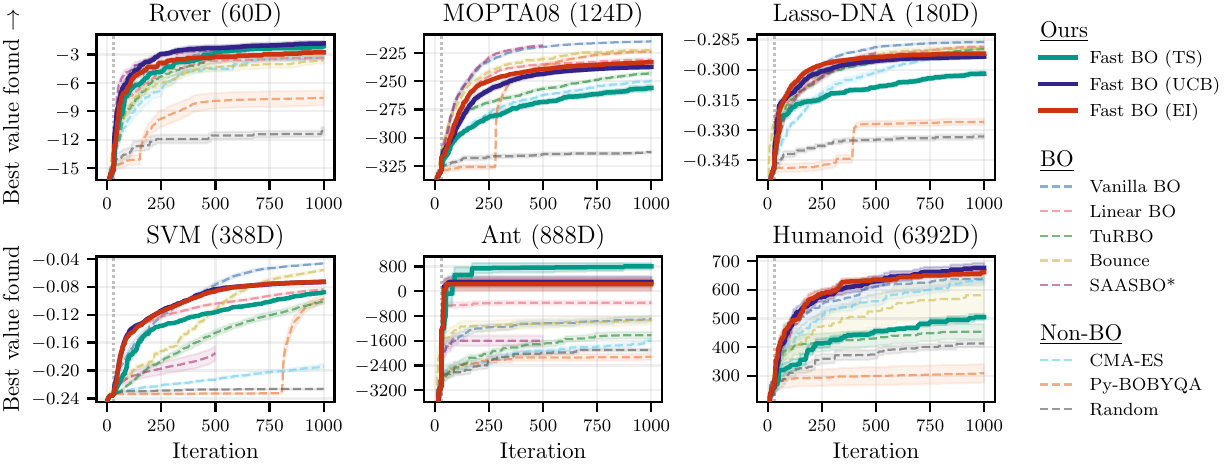}
  \caption{Extension of \Cref{fig:pareto-cont-high} over the course of the optimization.}
  \label{fig:sota}
\end{figure*}

\begin{figure*}[p]
  \centering
  \setlength{\abovecaptionskip}{3pt}
  \includegraphics[scale=0.806]{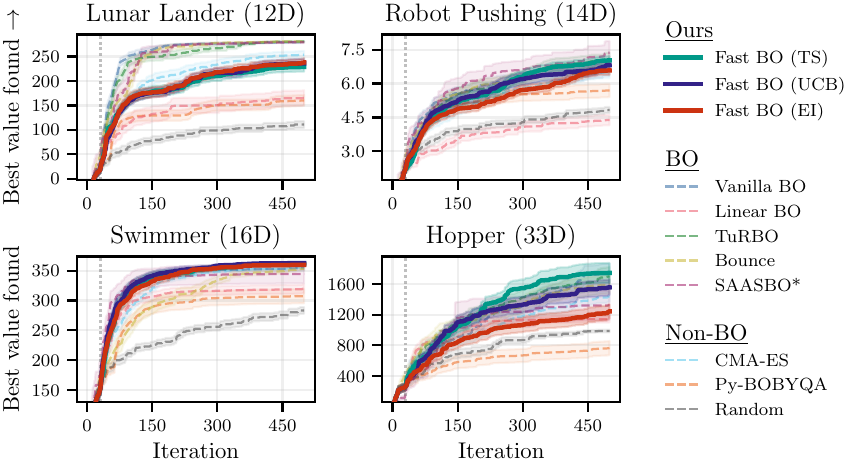}
  \caption{Extension of \Cref{fig:pareto-cont-low} over the course of the optimization.}
  \label{fig:sota-lowdim}
\end{figure*}

\begin{figure*}[p]
  \centering
  \setlength{\abovecaptionskip}{3pt}
  \includegraphics[scale=0.806]{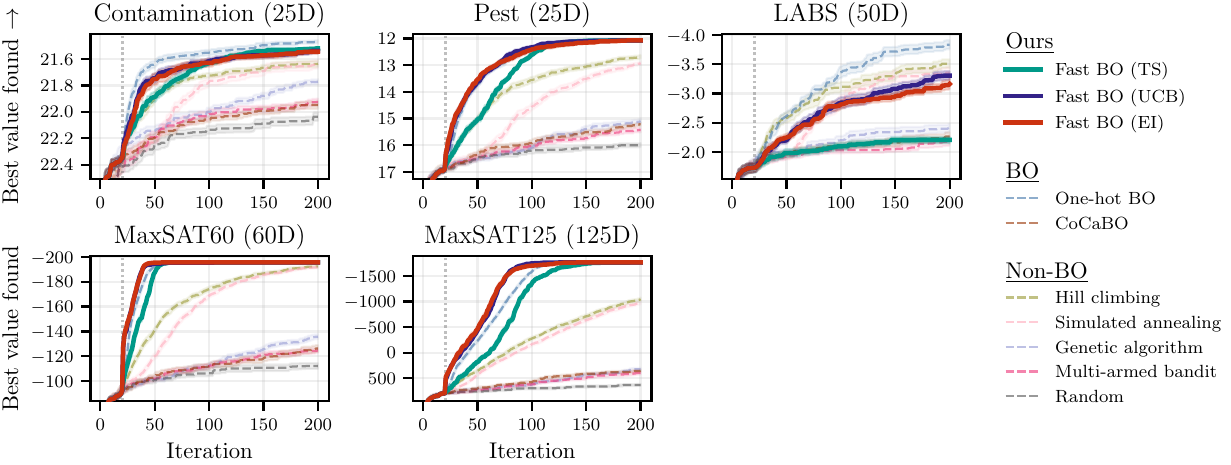}
  \caption{Extension of \Cref{fig:pareto-comb} over the course of the optimization.}
  \label{fig:sota-comb}
\end{figure*}

\Cref{fig:sota,fig:sota-lowdim,fig:sota-comb} show the best value found against the number of evaluations on the high-dimensional, low-dimensional, and combinatorial benchmarks, respectively. Each figure includes the methods from the corresponding Pareto figure, with our methods as solid lines and the baselines as dashed lines.

\subsection{Algorithm Time} \label{app:timing}

\Cref{tab:timing-full} gives the algorithm time per run on the high-dimensional benchmarks, displaying the speedup over Vanilla BO (with the wall-clock time in parentheses). We use Vanilla BO as the reference, given that it is currently the state of the art, as well as the only model-based baseline that runs on all six tasks on CPU. The speedup of our three AFs is largest on \texttt{Rover} and \texttt{Lasso-DNA} (over $10^3$), and smallest on \texttt{SVM} and \texttt{Humanoid} (a few hundred). Overall, our methods run in seconds to minutes, while Vanilla BO needs hours or even more than a day. Note that the spherical-linear model of \citet{doumont2026linear} does not by itself make Linear BO fast: Linear BO is significantly slower than Vanilla BO on \texttt{MOPTA08}, \texttt{SVM} and \texttt{Humanoid} ($0.51\times$, $0.14\times$ and $0.58\times$).

\begin{table*}[t]
\centering\footnotesize
\caption{Algorithm time per run on the high-dimensional benchmarks (model fitting and AF optimization, mean over $10$ seeds, excluding black-box evaluations). Each entry is the speedup over Vanilla BO (the dash), with the wall-clock time in gray parentheses. Benchmarks are sorted by dimension ($60$ to $6392$). \methodname runs on CPU throughout, whereas \texttt{TuRBO} and Linear BO use a GPU on \texttt{Humanoid} ($^\dagger$). These are the times plotted in \Cref{fig:pareto-cont-high}.}
\label{tab:timing-full}
\vspace{2pt}
\setlength{\tabcolsep}{4pt}
\begin{tabular}{@{}l l l l l l l@{}}
\toprule
& \multicolumn{3}{c}{\methodname} & \multicolumn{3}{c}{Baselines} \\
\cmidrule(r){2-4}\cmidrule(l){5-7}
Benchmark & TS & UCB & EI & \texttt{TuRBO} & Vanilla BO & Linear BO \\
\midrule
\texttt{Rover}     & $3254\times$ \tcell{$6.0$s} & $1366\times$ \tcell{$14$s} & $1251\times$ \tcell{$16$s} & $3.2\times$ \tcell{$1.7$h} & \textbf{--} \tcell{$5.4$h} & $7.3\times$ \tcell{$45$m} \\
\texttt{MOPTA08}   & $1747\times$ \tcell{$8.1$s} & $740\times$ \tcell{$19$s} & $690\times$ \tcell{$20$s} & $4.6\times$ \tcell{$51$m} & \textbf{--} \tcell{$3.9$h} & $0.51\times$ \tcell{$7.7$h} \\
\texttt{Lasso-DNA} & $2036\times$ \tcell{$15$s} & $1266\times$ \tcell{$24$s} & $1175\times$ \tcell{$26$s} & $11\times$ \tcell{$46$m} & \textbf{--} \tcell{$8.5$h} & $1.5\times$ \tcell{$5.8$h} \\
\texttt{SVM}       & $415\times$ \tcell{$39$s} & $330\times$ \tcell{$49$s} & $329\times$ \tcell{$49$s} & $4.3\times$ \tcell{$1.1$h} & \textbf{--} \tcell{$4.5$h} & $0.14\times$ \tcell{$31$h} \\
\texttt{Ant}       & $994\times$ \tcell{$1.6$m} & $812\times$ \tcell{$2.0$m} & $810\times$ \tcell{$2.0$m} & $2.4\times$ \tcell{$11$h} & \textbf{--} \tcell{$27$h} & $4.0\times$ \tcell{$6.7$h} \\
\texttt{Humanoid}  & $384\times$ \tcell{$4.1$m} & $306\times$ \tcell{$5.2$m} & $292\times$ \tcell{$5.4$m} & $7.3\times$ \tcell{$3.6$h$^\dagger$} & \textbf{--} \tcell{$26$h} & $0.58\times$ \tcell{$46$h$^\dagger$} \\
\bottomrule
\end{tabular}
\end{table*}

\begin{figure*}[p]
  \centering
  \setlength{\abovecaptionskip}{3pt}
  \includegraphics[scale=0.806]{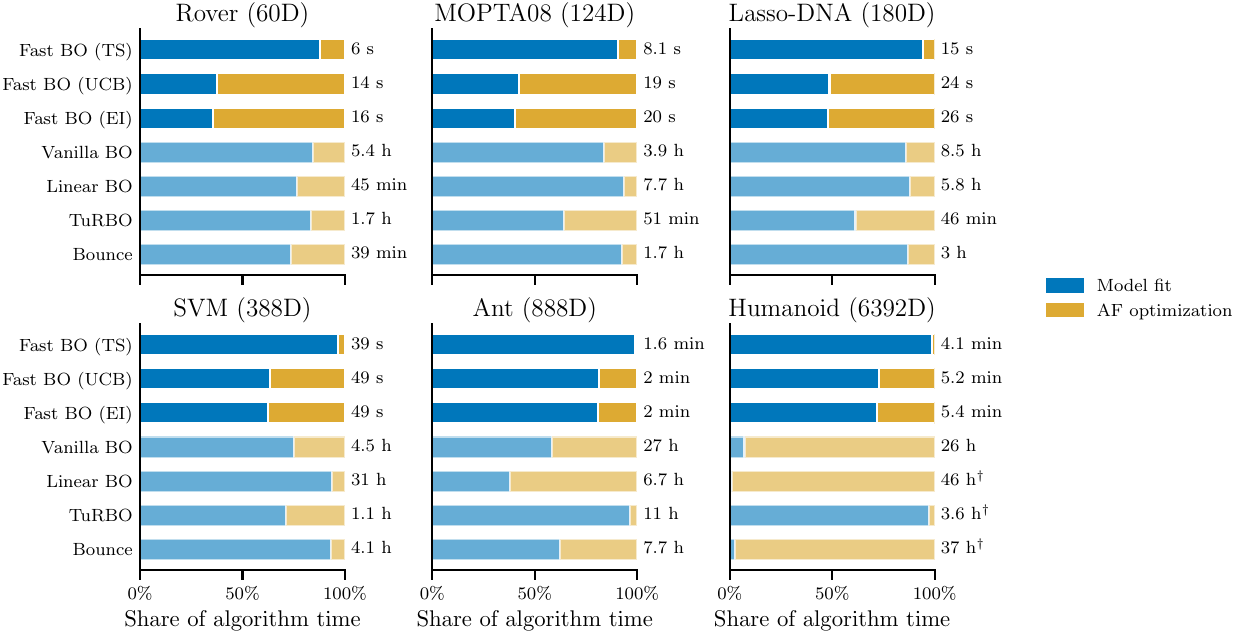}
  \caption{Depending on the problem, either the fit or the AF optimization can dominate the algorithm time of most BO baselines. In contrast, our TS method spends almost all of its time in the fit ($^\dagger$GPU).}
  \label{fig:time-breakdown}
\end{figure*}

\begin{figure*}[p]
  \centering
  \setlength{\abovecaptionskip}{3pt}
  \includegraphics[scale=0.806]{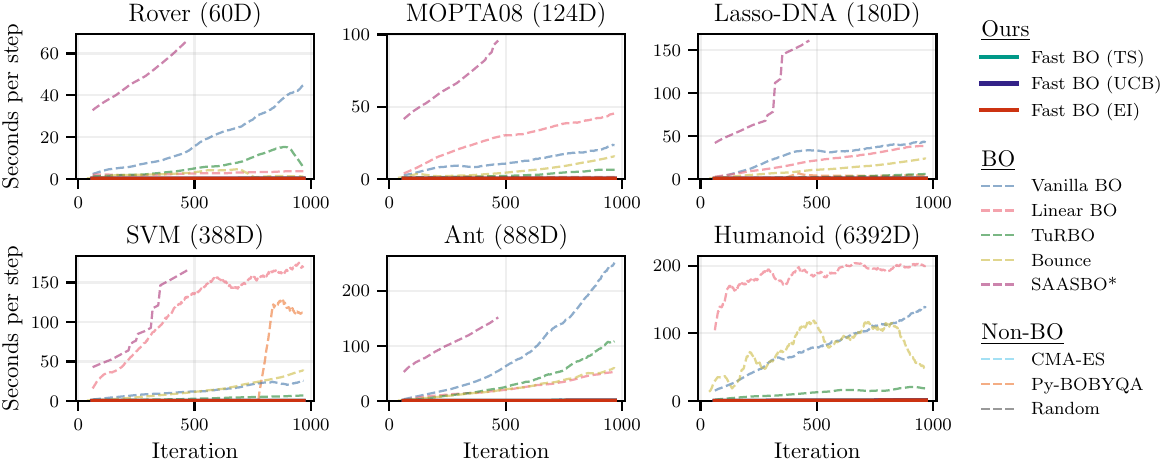}
  \caption{Our AFs take at most $5$ seconds per step, which is negligible compared to the BO baselines (median over seeds, smoothed with a $64$-step rolling mean).}
  \label{fig:time-per-step}
\end{figure*}

\Cref{fig:time-breakdown} splits the algorithm time of each run into fitting the model and optimizing the AF. For the baselines, this split depends on the problem. With a budget of $N = 1000$, fitting dominates when $N > D$ (\texttt{Rover} to \texttt{SVM}), both steps take a similar share when $N \approx D$ (\texttt{Ant}), and when $N < D$ (\texttt{Humanoid}), Vanilla BO, Linear BO and \texttt{Bounce} spend over $90\%$ of their time optimizing the AF. Both steps can thus be the bottleneck. For our methods, TS spends about $90$--$99\%$ of its time in the fit, since a single Dinkelbach solve is cheap, while UCB and EI spend roughly $20$--$65\%$ of their time in the AF. \Cref{fig:time-per-step} shows the time per step as the number of observations grows. The BO baselines take seconds to minutes per step, and slow down as $N$ grows. At these budgets, their cost per step grows roughly linearly to quadratically in $N$, as computing the kernel and its gradients ($O(N^2D)$) dominates the $O(N^3)$ Cholesky decomposition. Our methods take from tens of milliseconds (\texttt{Rover}) to a few seconds (\texttt{Humanoid}) per step.

\subsection{Ablations} \label{app:ablation}

\begin{figure*}[p]
  \centering
  \setlength{\abovecaptionskip}{3pt}
  \includegraphics[scale=0.806]{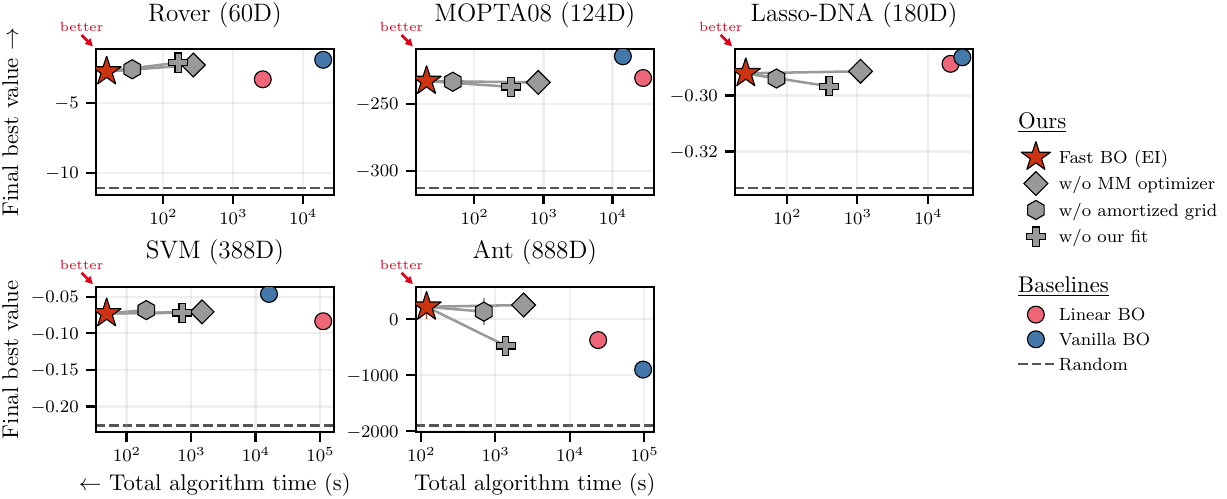}
  \caption{Each ingredient of Fast BO (EI) contributes to its speed, with our MM optimizer contributing most (about $20$--$40\times$). Only the gradient-based fit lowers the final value on some benchmarks.}
  \label{fig:ablation-loo}
\end{figure*}

\paragraph{High-Dimensional.}
\Cref{fig:ablation-loo} removes one ingredient of Fast BO (EI) at a time, on all high-dimensional benchmarks except \texttt{Humanoid}. First, replacing our MM optimizer by the LogEI \citep{ament2023logei} of \texttt{BoTorch}, maximized with the \texttt{optimize\_acqf} settings of Linear BO, makes a run approximately $20$--$40\times$ slower. Second, running the full lengthscale grid at every fit, instead of every tenth fit, makes it roughly $2$--$6\times$ slower. Third, fitting the same shared-lengthscale model by gradient-based MAP, instead of our grid search and Nelder--Mead, makes it about $10$--$20\times$ slower. The first two changes leave the final value mostly within noise, while the gradient-based fit is worse on \texttt{MOPTA08}, \texttt{Lasso-DNA} and \texttt{Ant}, and better on \texttt{Rover}. We do not ablate the shared lengthscale, since our grid search is only possible with a single lengthscale. For reference, we also show Linear BO, which has none of the three ingredients.

\begin{figure*}[p]
  \centering
  \setlength{\abovecaptionskip}{3pt}
  \includegraphics[scale=0.806]{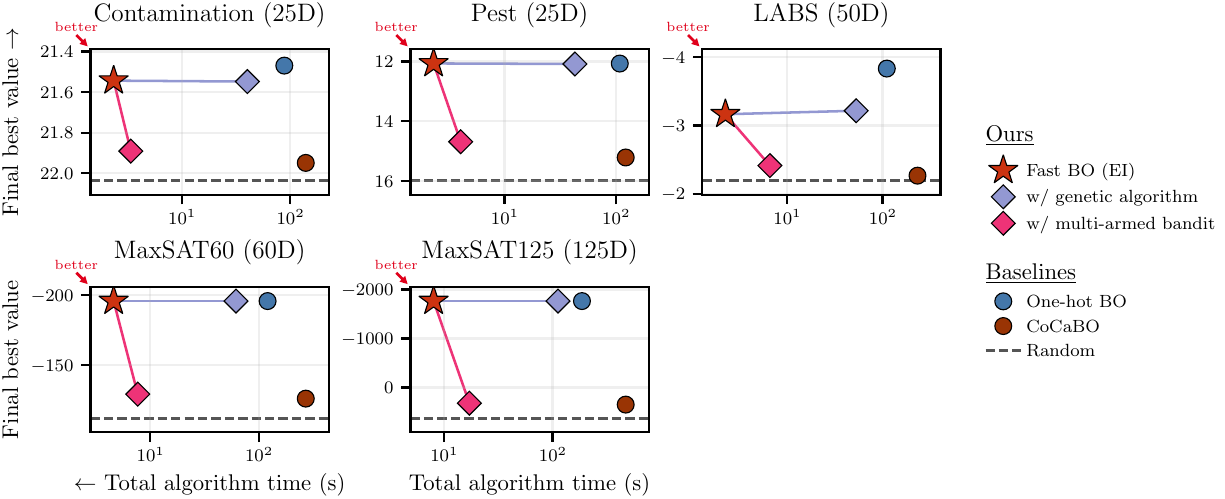}
  \caption{The genetic algorithm reaches the same values as our MM optimizer, but is about $10$--$20\times$ slower. The multi-armed bandit drops to the level of \texttt{CoCaBO}. (The $y$ axis is inverted since the tasks are minimized.)}
  \label{fig:ablation-comb}
\end{figure*}

\paragraph{Combinatorial.}
\Cref{fig:ablation-comb} keeps our combinatorial model and EI, but maximizes EI with the genetic algorithm or the multi-armed bandit of \texttt{MCBO}, instead of our MM optimizer. The genetic algorithm reaches the same final values, but is approximately $10$--$20\times$ slower. With the multi-armed bandit, performance drops to the level of \texttt{CoCaBO}, which combines the same optimizer with a GP. This suggests that the weaker results of \texttt{CoCaBO} are due to its AF optimizer rather than its kernel.

\begin{figure*}[t]
  \centering
  \setlength{\abovecaptionskip}{3pt}
  \includegraphics[scale=0.806]{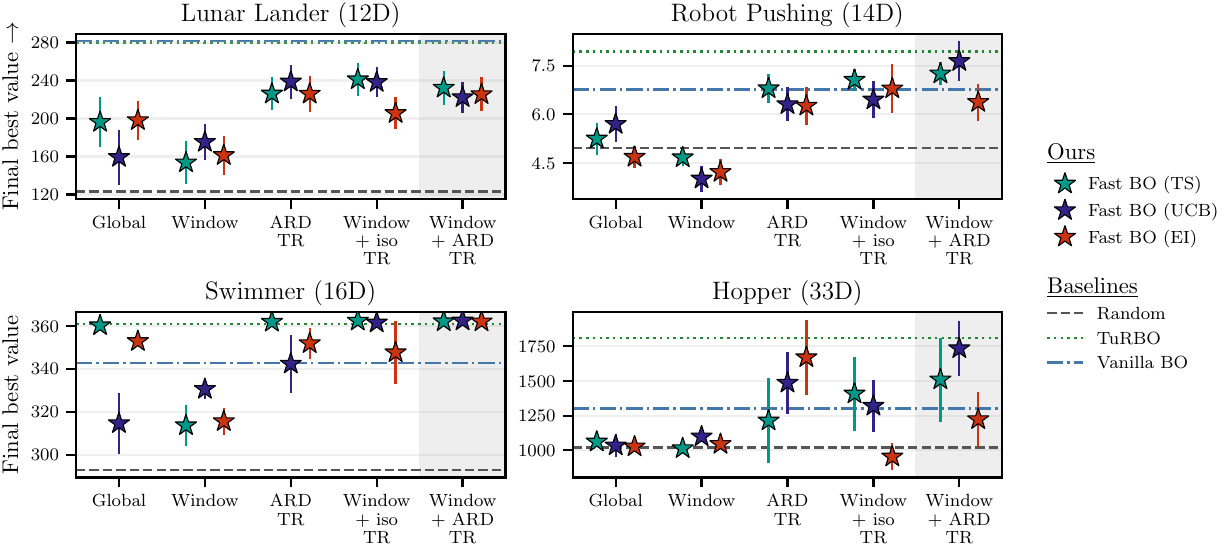}
  \caption{Most of the improvement of our low-dimensional variant is due to the ARD trust region, whereas the window mainly helps once combined with this trust region (mean over the first $10$ of the $40$ seeds).}
  \label{fig:ablation-lowdim}
\end{figure*}

\paragraph{Low-Dimensional.} \label{app:ablation-lowdim}
\Cref{fig:ablation-lowdim} isolates the effects of the trust region and the window fit from \Cref{app:tr} on the four low-dimensional benchmarks, for each of our AFs. Most of the improvement is due to the trust region, which improves upon plain Fast BO in nearly all cases. In contrast, the window alone does not help, and often degrades performance. When combined with the trust region, however, it improves the results in most cases. Lastly, an isotropic trust region combined with the window is faster than our method, but performs slightly worse. Due to its cost, this ablation only uses the first $10$ of the $40$ seeds (for all variants and reference lines). Its values therefore differ from \Cref{fig:pareto-cont-low}, most visibly on \texttt{Hopper}, where the variance across seeds is high.

\end{document}